%% file: main.tex
\documentclass[10pt,twocolumn,letterpaper]{article}

\usepackage[pagenumbers]{cvpr} 

\input{preamble}

\definecolor{cvprblue}{rgb}{0.21,0.49,0.74}
\usepackage[pagebackref,breaklinks,colorlinks,allcolors=cvprblue]{hyperref}

\def\confName{CVPR}
\def\confYear{2027}

\title{\wgaze: Do Vision Large Language Models Know\\ Who Looks at What, and When?}

\author{Shayan Nasiriboukani$^{1}$, Mohammadreza Taheriboukani, Wish Suharitdamrong$^{2}$, Semco Jahanbin,\\
Sara Atito$^{1,2,\star}$, Muhammad Awais$^{1,2,\star}$\\[3pt]
\small $^{1}$\,Centre for Vision, Speech and Signal Processing (CVSSP), University of Surrey, UK\\
\small $^{2}$\,Surrey Institute for People-Centred AI, University of Surrey, UK\\[2pt]
{\tt\small \{s.nasiriboukani,sara.atito,muhammad.awais\}@surrey.ac.uk}\\[-16pt]
}

\begin{document}
\maketitle
\begingroup\renewcommand\thefootnote{}\footnotetext{$^\star$Corresponding authors.}\endgroup

\begin{strip}
    \centering
    \includegraphics[width=\textwidth]{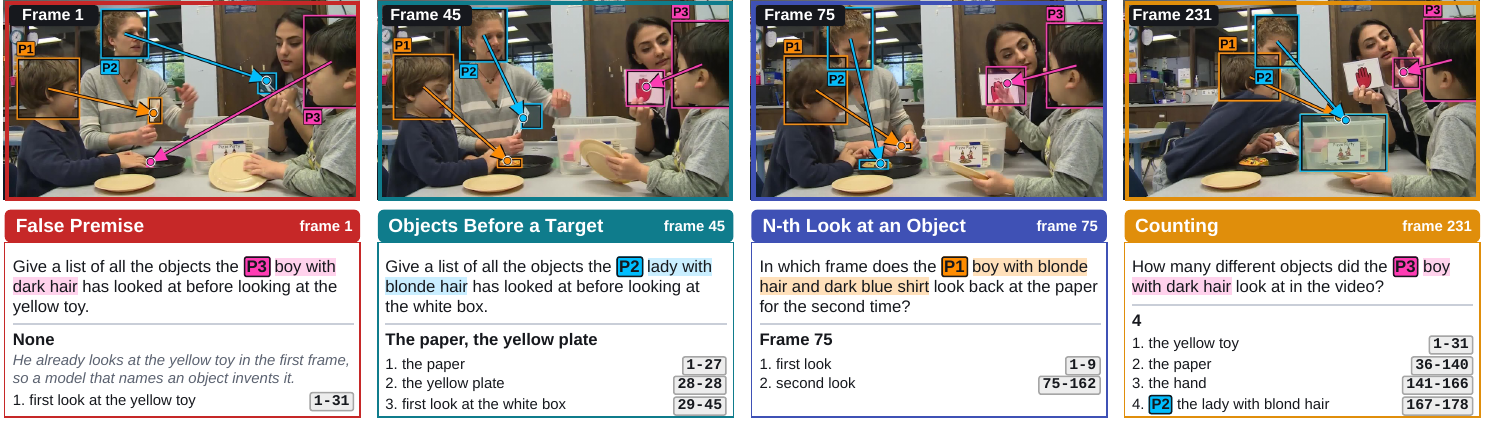}
    \captionof{figure}{\textbf{Temporal questions in \wgaze.}
    Each question is asked on a whole video, and its card lists the
    answer frame, the answer, and the frame
    ranges that the answer comes from. Colours mark the same
    person in the frames and in the text.}
    \label{fig:gaze_vqa_examples}
\end{strip}

\input{sec/0_abstract}    
\input{sec/1_intro}
\input{sec/2_related_work}

\input{sec/3_benchmark}
\input{sec/4_experiments}

\input{sec/6_conclusion}
\par\flushbottom
{
    \small
    \bibliographystyle{ieeenat_fullname}
    \bibliography{main}
}

\clearpage
\setcounter{section}{0}\setcounter{table}{0}\setcounter{figure}{0}\setcounter{equation}{0}
\renewcommand{\thesection}{S\arabic{section}}\renewcommand{\theHsection}{supp.\arabic{section}}
\renewcommand{\thetable}{S\arabic{table}}\renewcommand{\theHtable}{supp.\arabic{table}}
\renewcommand{\thefigure}{S\arabic{figure}}\renewcommand{\theHfigure}{supp.\arabic{figure}}
\renewcommand{\theequation}{S\arabic{equation}}
\maketitlesupplementary
\definecolor{cSub}{HTML}{EEEEEE}   

This supplement collects the details that the main paper leaves out. \Cref{sec:supp-config} lists the settings of the VLLM evaluation and of our gaze tracking agent, together with the results of all gaze estimators that we tested. \Cref{sec:supp-prompts} shows the prompts that the VLLMs and the judge receive, with the complete input of three example questions. \Cref{sec:supp-examples} then shows example videos for each question type of \wgaze, two for almost every type, so that the quality of the questions and their labels can be judged directly. \Cref{sec:supp-ablation} finally changes only the input of the VLLMs, from the question alone up to the annotated gaze, to show which questions a gaze arrow helps.

\section{Implementation Details and Configuration}
\label{sec:supp-config}

The main paper describes the evaluation at the level needed to follow the results, and this section collects the settings behind it. The paragraphs below give the settings of the VLLM evaluation, and \cref{tab:supp-agent} lists those of the gaze tracking agent. All values were read from the configuration files and the job logs of the final runs.

\paragraph{Serving and decoding}
Every VLLM is served with vLLM 0.20.1 on NVIDIA H100 GPUs, and each request holds one question with one image or one video. We decode greedily with a temperature of 0 and allow up to 4{,}096 new tokens. Thinking is switched off for every model, because with thinking on the Qwen3.5 models spend the token budget on reasoning before they reach an answer, and no answer format is enforced. All models run in BF16, except Qwen3.5-122B-A10B, which runs from its FP8 checkpoint.

\paragraph{Visual input}
Frame-level questions are asked on the full frame, with a blue arrow from the centre of the head box of each relevant person to the gaze point that our agent predicts with PaGE~\cite{ye2026page}. The arrows are 5 px wide, the width that worked best in the visual prompt ablation of GTA~\cite{nasiriboukani2026gaze}. A question about one person shows only that person's arrow, a scene question shows the arrows of all people, and an ambiguous description shows no arrow, so that the arrow cannot reveal the person. Temporal questions are asked on a video of the whole clip with at most 64 frames, and every frame carries its clip frame number in the top left corner, because the answers refer to frame numbers. When a clip has more than 64 frames, the sampled frames always include the answer and evidence frames, so a frame answer can always be read from the video.

\begin{table}[t]
\centering
\caption{\textbf{Gaze tracking agent settings.} The agent produces the gaze arrows that the VLLMs see.}
\footnotesize
\renewcommand{\arraystretch}{1.05}
\setlength{\tabcolsep}{4pt}
\begin{tabular}{p{0.36\columnwidth}p{0.54\columnwidth}}
\toprule
\textbf{Setting} & \textbf{Value} \\
\midrule
\rowcolor{cSub}\multicolumn{2}{l}{\textbf{Gaze tracking agent (\cref{tab:gaze_tracking_all})}} \\
Gaze estimator & PaGE, ViT-B \\
Head and eye monitor & OpenFace 3.0 \\
Trigger threshold & $2.0^{\circ}$ change of head or eye direction between two frames \\
Trigger spacing & One trigger per run of changes, at least 3 frames apart \\
Drift trigger & Point more than 1.0 head diagonal away from the last estimate \\
In/out threshold & 0.35 \\
Point tracker & Pyramidal Lucas--Kanade, $21{\times}21$ px window, 3 levels \\
Stopping rule & 30 iterations or a change below 0.01 \\
Forward backward check & 8 px \\
Search window & 30 px radius, up to 8 corners \\
\bottomrule
\end{tabular}

\label{tab:supp-agent}
\end{table}

\paragraph{Gaze tracking agent}
The agent runs the gaze estimator only when a new target has to be found. OpenFace 3.0~\cite{hu2025openface30lightweightmultitask} gives the head and eye direction of every person in every frame, and a change of more than $2^{\circ}$ between two frames starts a trigger. Several such frames in a row form one change, and only its peak frame calls the estimator, with at least 3 frames between two calls. Between calls, the gaze point is followed with pyramidal Lucas--Kanade optical flow on up to 8 corners found within 30 px of the point, and a corner is kept only when tracking it back to the previous frame lands within 8 px of where it started. If the tracked point moves more than one head diagonal away from the last estimate, the estimator is called again. A person counts as looking outside the frame when the in/out score of the estimator is below 0.35.

\subsection{All gaze estimators}
\label{sec:supp-estimators}

\Cref{tab:gaze_tracking_all} extends \cref{tab:gaze_tracking_comparison} of the main paper to all 14 gaze estimators that we tested, once on every frame of every person and once inside our tracking agent.

\begin{figure*}[t]
\centering
\includegraphics[width=\textwidth]{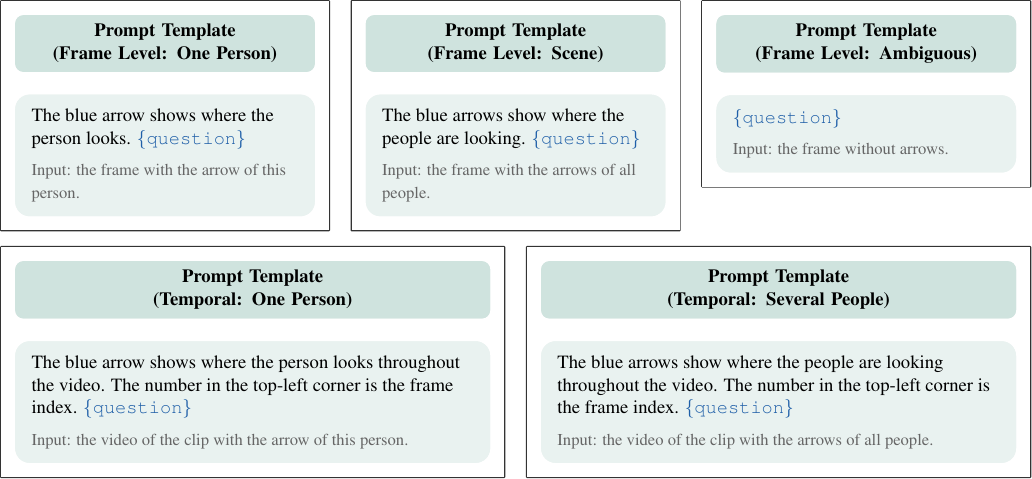}
\caption{\textbf{VLLM prompt templates.} One sentence before the benchmark question explains the blue gaze arrows. Which template is used depends on whether the question is about one person, about the scene, or about an ambiguous description, and on whether it is asked on a frame or on the whole clip. The same templates are used for all 16 VLLMs.}
\label{fig:supp-vllm-prompts}
\end{figure*}

\begin{table}[ht]
\centering
\caption{\textbf{Gaze tracking accuracy and efficiency of all gaze estimators.} {\setlength{\fboxsep}{1pt}\colorbox{rank1}{Red}} marks the best result, {\setlength{\fboxsep}{1pt}\colorbox{rank2}{orange}} the second, and {\setlength{\fboxsep}{1pt}\colorbox{rank3}{yellow}} the third.}
\label{tab:gaze_tracking_all}

\scriptsize
\setlength{\tabcolsep}{3pt}
\renewcommand{\arraystretch}{1.08}

\resizebox{\columnwidth}{!}{%
\begin{tabular}{@{}lccccc@{}}
\toprule
\textbf{Method} & \textbf{Calls $\downarrow$} & \textbf{FLOPs/Call} & \textbf{AUC $\uparrow$} & \textbf{Avg. L2 $\downarrow$} & \textbf{In/out AP $\uparrow$} \\
\midrule
PaGE & 25,017 & 0.293 T & \cellcolor{rank1}0.9709 & \cellcolor{rank1}0.0730 & \cellcolor{rank1}0.9980 \\
Gaze-LLE & 25,017 & 0.733 T & \cellcolor{rank3}0.9493 & \cellcolor{rank3}0.1016 & 0.9916 \\
GazeAnywhere & 25,017 & 0.922 T & 0.9416 & 0.1190 & 0.9658 \\
Sharingan & 25,017 & \cellcolor{rank2}0.048 T & 0.9258 & 0.1134 & 0.9905 \\
Gupta et al. & 25,017 & 1.392 T & 0.9218 & 0.1378 & -- \\
PDP & 25,017 & 0.964 T & 0.9199 & 0.1248 & 0.9918 \\
MTGS & 25,017 & 0.334 T & 0.9193 & 0.1039 & \cellcolor{rank3}0.9955 \\
GeomGaze & 25,017 & 2.595 T & 0.9167 & 0.1175 & 0.9842 \\
SemGaze & 25,017 & \cellcolor{rank3}0.056 T & 0.9163 & 0.1124 & -- \\
Chong et al. & 25,017 & \cellcolor{rank1}0.018 T & 0.9083 & 0.1304 & 0.9529 \\
Tonini et al. & 25,017 & 0.973 T & 0.9039 & 0.1244 & 0.9491 \\
ViTGaze & 25,017 & 0.094 T & 0.9031 & 0.1125 & 0.9904 \\
GOTD & 25,017 & 0.193 T & 0.7456 & 0.2323 & 0.9838 \\
GazeHTA & 25,017 & 1.903 T & 0.7242 & 0.2303 & 0.9275 \\
\midrule
Ours w/ PaGE & 3,816 & 0.293 T & \cellcolor{rank2}0.9705 & \cellcolor{rank2}0.0807 & \cellcolor{rank2}0.9979 \\
Ours w/ Gaze-LLE & 3,818 & 0.733 T & 0.9485 & 0.1078 & 0.9909 \\
Ours w/ GazeAnywhere & 3,819 & 0.922 T & 0.9430 & 0.1219 & 0.9637 \\
Ours w/ Sharingan & 3,816 & \cellcolor{rank2}0.048 T & 0.9241 & 0.1185 & 0.9891 \\
Ours w/ Gupta et al. & 3,818 & 1.392 T & 0.9187 & 0.1432 & -- \\
Ours w/ PDP & 3,815 & 0.964 T & 0.9183 & 0.1305 & 0.9898 \\
Ours w/ MTGS & \cellcolor{rank2}3,813 & 0.334 T & 0.9177 & 0.1087 & 0.9945 \\
Ours w/ GeomGaze & \cellcolor{rank2}3,813 & 2.595 T & 0.9166 & 0.1192 & 0.9800 \\
Ours w/ SemGaze & 3,815 & \cellcolor{rank3}0.056 T & 0.9134 & 0.1192 & -- \\
Ours w/ Chong et al. & \cellcolor{rank1}3,812 & \cellcolor{rank1}0.018 T & 0.9062 & 0.1333 & 0.9508 \\
Ours w/ Tonini et al. & \cellcolor{rank1}3,812 & 0.973 T & 0.9015 & 0.1283 & 0.9488 \\
Ours w/ ViTGaze & 3,817 & 0.094 T & 0.9019 & 0.1182 & 0.9900 \\
Ours w/ GOTD & \cellcolor{rank3}3,814 & 0.193 T & 0.7457 & 0.2257 & 0.9833 \\
Ours w/ GazeHTA & \cellcolor{rank3}3,814 & 1.903 T & 0.7306 & 0.2276 & 0.9251 \\
\bottomrule
\end{tabular}%
}
\end{table}

\section{Prompt Templates}
\label{sec:supp-prompts}

Our evaluation uses two kinds of prompts in sequence. The first asks a VLLM the benchmark question, and the second asks the judge whether the answer of the VLLM is correct. Both are shown here word for word, with the parts that change between questions in blue braces.

\subsection{VLLM prompts}
\label{sec:supp-vllm-prompts}

\begin{figure*}[t]
\centering
\includegraphics[width=\textwidth]{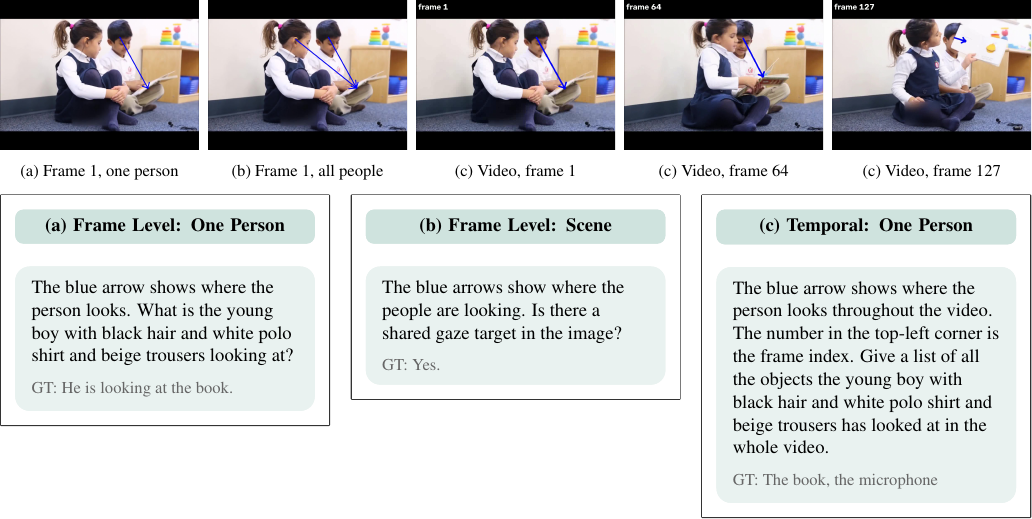}
\caption{\textbf{VLLM inputs for one clip.} PaGE-predicted gaze arrows show only the boy in (a) and both children in (b). The video (c) contains 44 numbered frames, three of which are shown. The GT answers are for the reader and are not given to the models.}
\label{fig:supp-vllm-inputs}
\end{figure*}

Every VLLM receives one user message with the image or the video and one short text, and no system message. The text is the benchmark question with a single sentence in front of it that explains the blue arrows (\cref{fig:supp-vllm-prompts}). Questions about one person use the singular sentence, because the input shows only that person's arrow, and questions about the scene use the plural sentence, because the input shows the arrows of all people. Ambiguous descriptions get the question alone, since their input has no arrow. The temporal prompts add a second sentence that points the model to the frame number in the corner of every frame, because the answers to temporal questions are frame numbers and not seconds. We do not ask for any answer format, so the answers are free text, which are checked by exact match or by the judge of \cref{sec:supp-judge-prompt}. \Cref{fig:supp-vllm-inputs} shows the complete input for three questions about the same clip.

\subsection{Judge prompt}
\label{sec:supp-judge-prompt}

\begin{figure}[t]
\centering
\begin{minipage}{0.9\columnwidth}
\centering
\includegraphics[width=\linewidth]{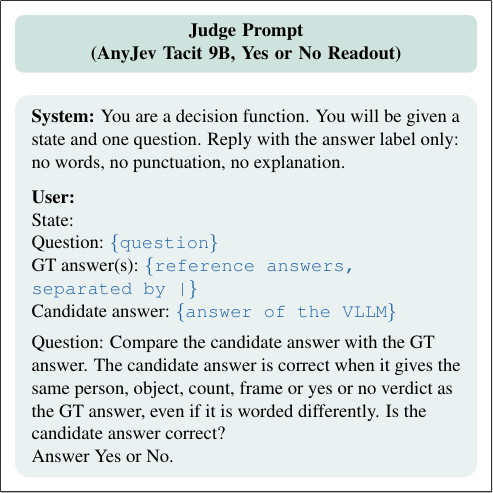}
\caption{\textbf{Judge prompt.} The judge sees text only and the score is the probability of the label \emph{Yes}.}
\label{fig:supp-judge-prompt}
\end{minipage}
\end{figure}

Answers that cannot be checked by exact match are scored by the AnyJev Tacit 9B judge (\cref{fig:supp-judge-prompt}). The judge receives a fixed system message and a state that holds the question, the reference answers and the candidate answer, followed by the judging question. It does not generate text. We read the probabilities of the two answer labels \emph{Yes} and \emph{No} from a single forward pass and use the probability of \emph{Yes} as the score of the answer, which counts as correct when it is at least 0.5. The wording asks whether the candidate gives the same person, object, count, frame, or verdict as the reference, so that a correct answer with different wording is not penalised.

\section{Examples of the Question Types}
\label{sec:supp-examples}

The following pages show two videos for each question type of \wgaze, first for the frame-level types and then for the 21 temporal types. The two videos of a type come from different source videos. For two temporal types (\cref{fig:supp-t09_person_object_list,fig:supp-t21_occluded_gaze_reasoning}) only one example passed all of our checks, so these types show one video. Every example was checked against the labels of the benchmark and by eye on its frames, and examples with an unclear target, a description that does not fit the person, or unclear wording were left out. Each video is drawn as a player with four frames in time order, numbered in the top left corner and at the same places on the progress bar. The questions and answers are copied from the benchmark without changes.

The arrows show the annotated gaze of the people that the question is about, so that the reader can check the answers. The VLLMs see the arrows of our tracking agent instead (\cref{fig:supp-vllm-inputs}). For frame-level questions the answer of every frame is printed under it. For temporal questions the answer and its GT looks are printed under the player, and each frame carries its clip frame number.

\subsection{Frame-level questions}
\label{sec:supp-frame-examples}

\begin{figure*}[p]
\centering
\includegraphics[width=\textwidth]{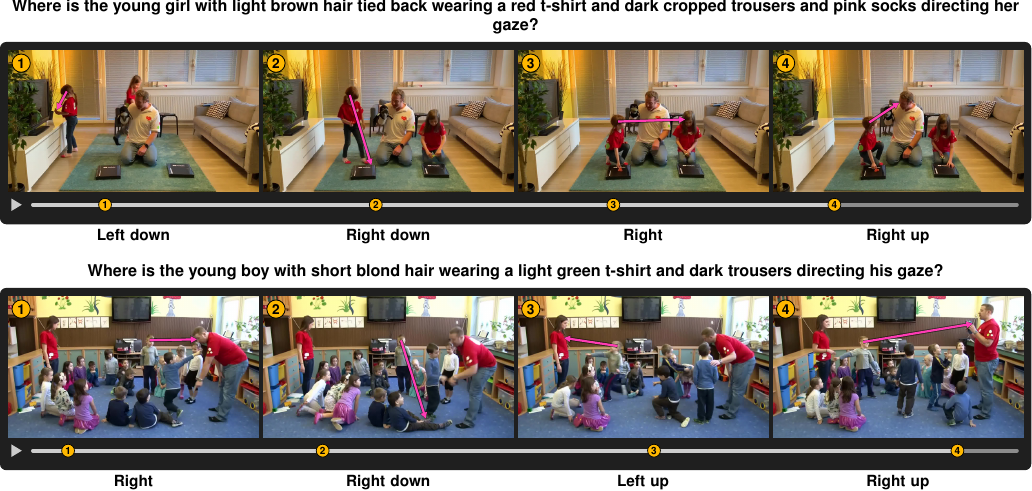}
\caption{\textbf{Gaze direction.} The question asks in which direction one person looks. The benchmark answer is a sentence such as \emph{He directs his gaze to the right down}, and only its direction is printed under each frame. These questions are answered by the gaze tracker and evaluate the tracking itself.}
\label{fig:supp-f02_gaze_direction}
\vspace{14pt}
\includegraphics[width=\textwidth]{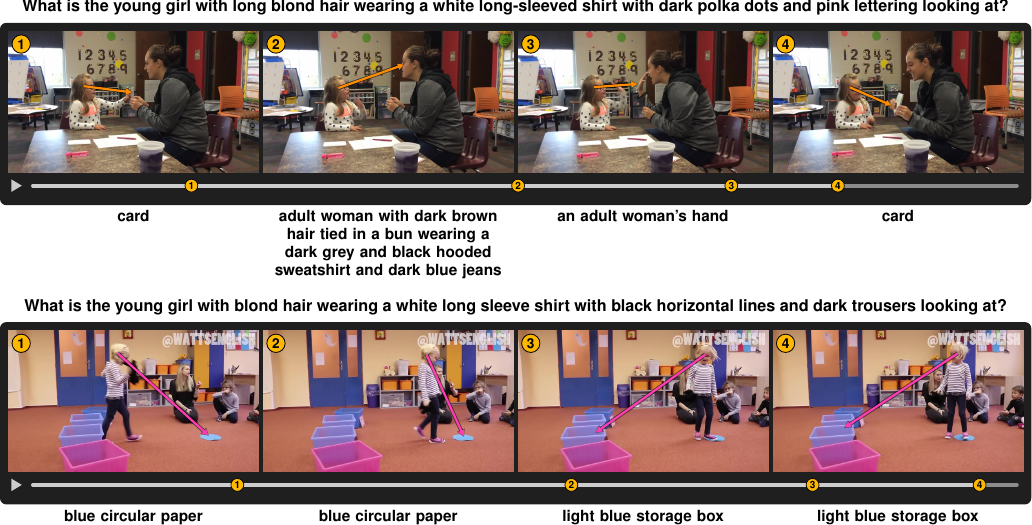}
\caption{\textbf{Gaze object description.} The person is named by appearance and the answer is the object or person that they look at, written in the benchmark as \emph{He is looking at the book}. Only the target is printed under each frame. The target changes within one clip, so the VLLM has to follow the arrow in every frame instead of naming an object that people often look at.}
\label{fig:supp-f03_gaze_object}
\end{figure*}

\begin{figure*}[p]
\centering
\includegraphics[width=\textwidth]{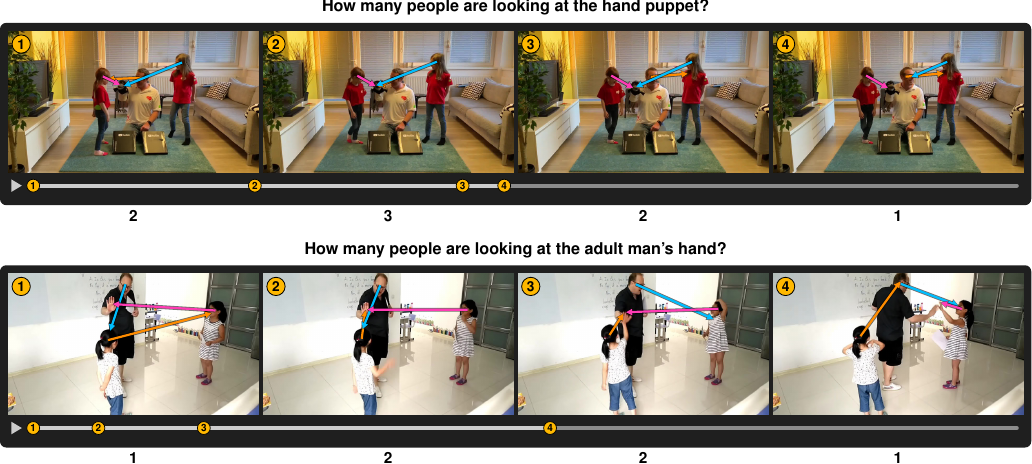}
\caption{\textbf{Counting.} The question asks how many people look at a named object or person, and the arrows of all people are drawn. The count changes over the clip as people turn to the object and away from it.}
\label{fig:supp-f04_counting}
\vspace{14pt}
\includegraphics[width=\textwidth]{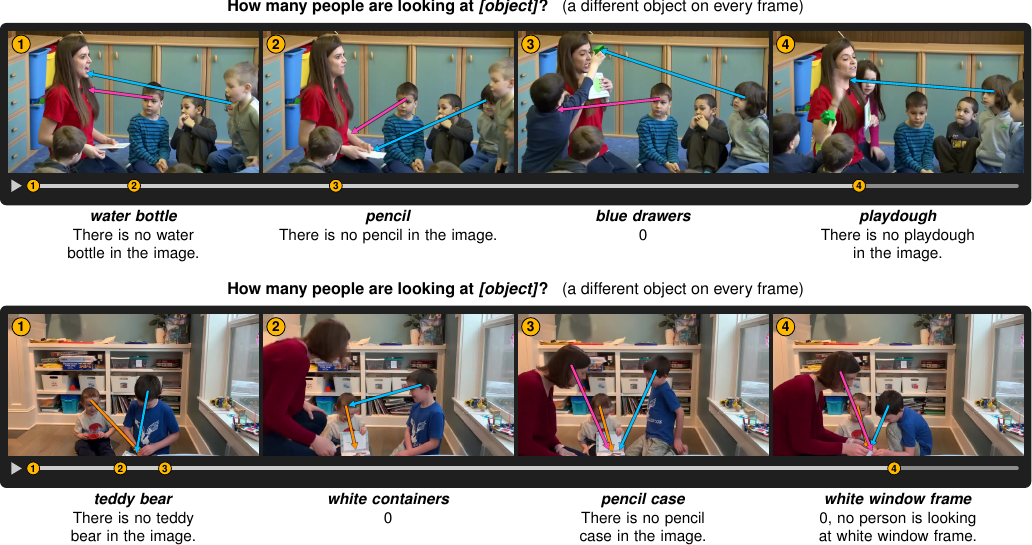}
\caption{\textbf{Absent object.} Every frame carries a question about a different object, printed in italics under the frame. The object is either not in the video, with an answer such as \emph{There is no ... in the image}, or visible but not looked at by anyone, with the answer 0. A VLLM that names someone who looks at the object is hallucinating.}
\label{fig:supp-f05_absent}
\end{figure*}

\begin{figure*}[p]
\centering
\includegraphics[width=\textwidth]{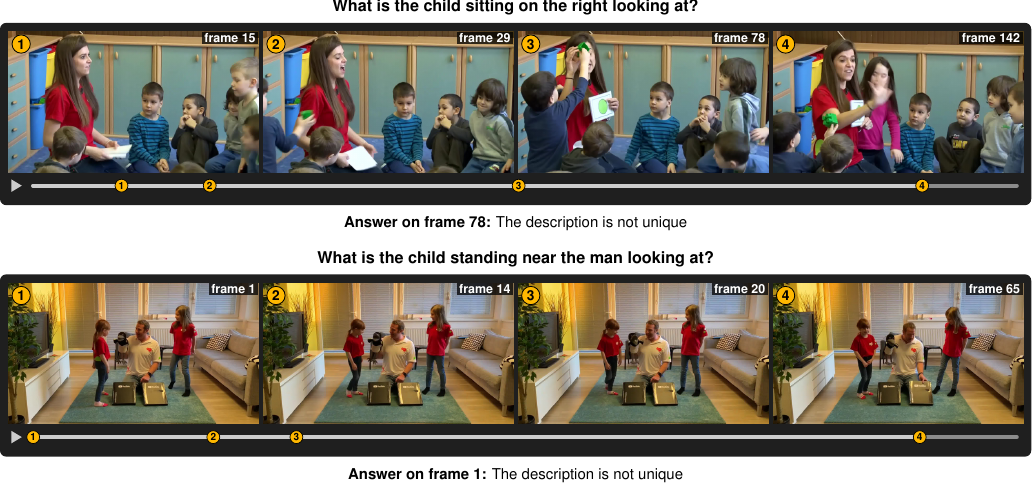}
\caption{\textbf{Ambiguous description.} The description in the question fits more than one person, so the correct answer is that the description is not unique. These questions are written by the annotators on key frames, and the answer names the frame on which the question was asked. As in the benchmark, no arrow is drawn, because an arrow would point out one person.}
\label{fig:supp-f06_ambiguous}
\vspace{14pt}
\includegraphics[width=\textwidth]{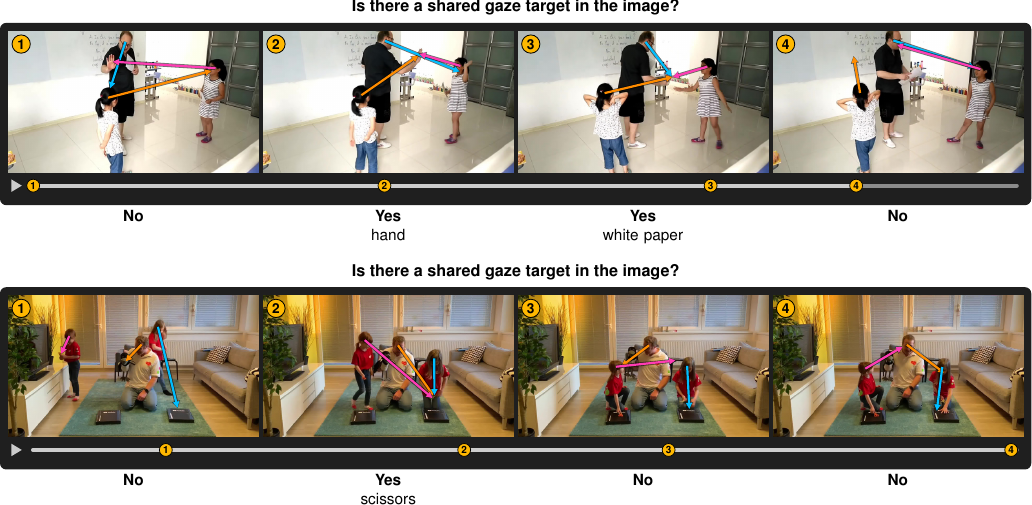}
\caption{\textbf{Shared gaze.} The question asks whether several people look at the same target, and the arrows of all people are drawn. Under each \emph{Yes} we print the answer to the question \emph{What is the main object people are looking at?} that is asked on the same frame.}
\label{fig:supp-f07_shared}
\end{figure*}

\Cref{fig:supp-f02_gaze_direction} to \cref{fig:supp-f08_exchanged} show seven of the nine frame-level types in the order of the main paper: gaze direction, then the attended object types, then shared and exchanged gaze.

\subsection{Temporal questions}
\label{sec:supp-temporal-examples}

\begin{figure*}[p]
\centering
\includegraphics[width=\textwidth]{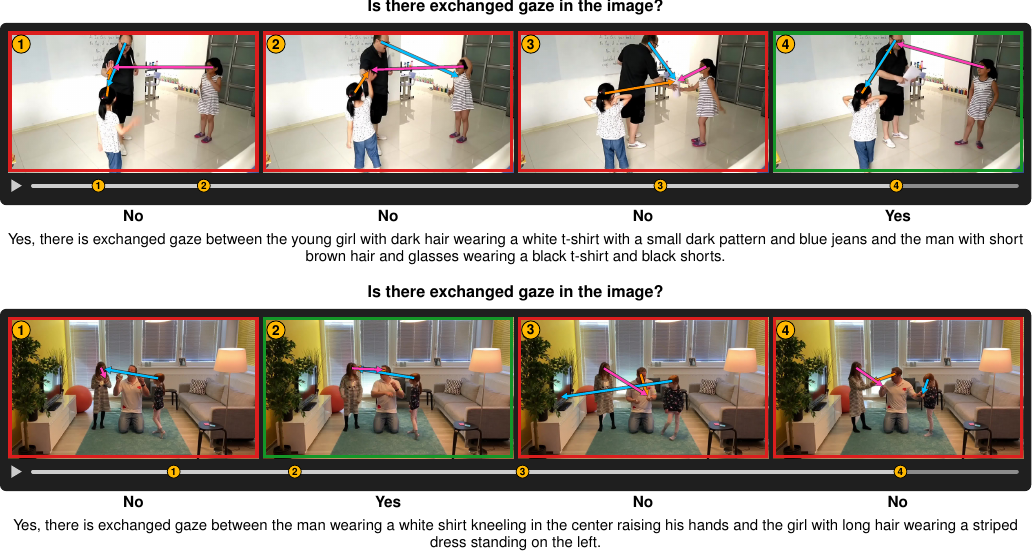}
\caption{\textbf{Exchanged gaze.} The question asks whether two people look at each other. The answer of every frame is printed under it, and the full answer of the first \emph{Yes} frame, which names both people, is printed under the player.}
\label{fig:supp-f08_exchanged}
\vspace{14pt}
\includegraphics[width=\textwidth]{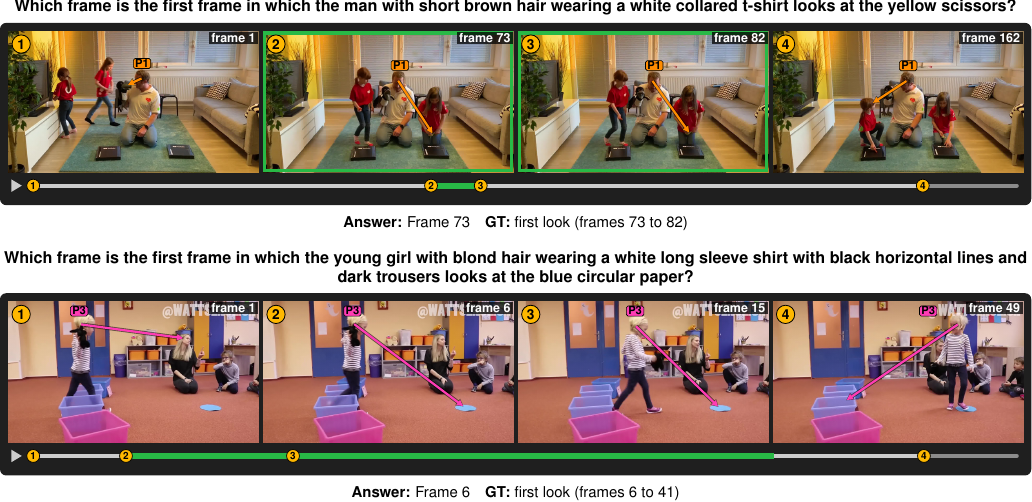}
\caption{\textbf{First frame of a look.} The question asks for the first frame in which a person, named by appearance, looks at an object. The answer is a frame number and the GT gives the frames of this first look.}
\label{fig:supp-t01_first_frame_person_object}
\end{figure*}

\begin{figure*}[p]
\centering
\includegraphics[width=\textwidth]{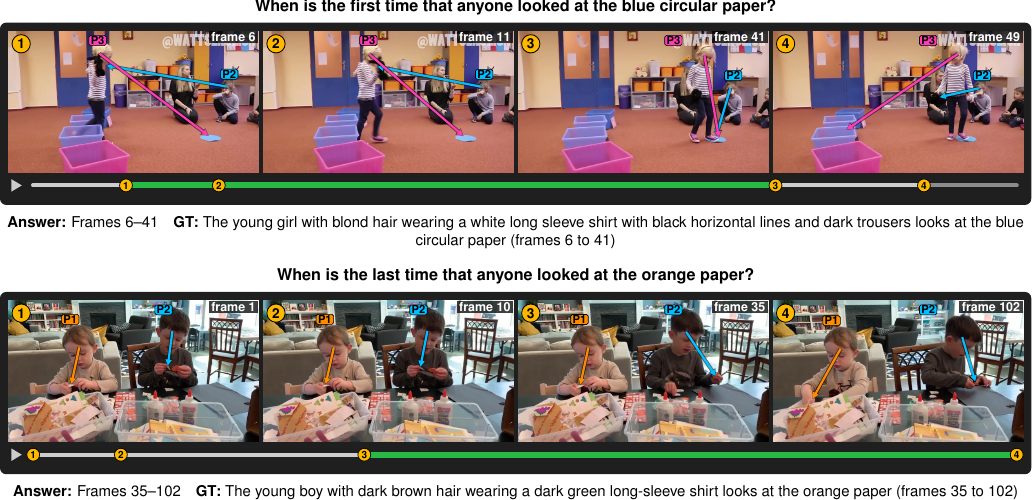}
\caption{\textbf{First or last look at an object.} The question asks when anyone looks at an object for the first or the last time. The answer is a range of frames, and the GT lists who looks at the object and when.}
\label{fig:supp-t02_first_last_look_at_object}
\vspace{14pt}
\includegraphics[width=\textwidth]{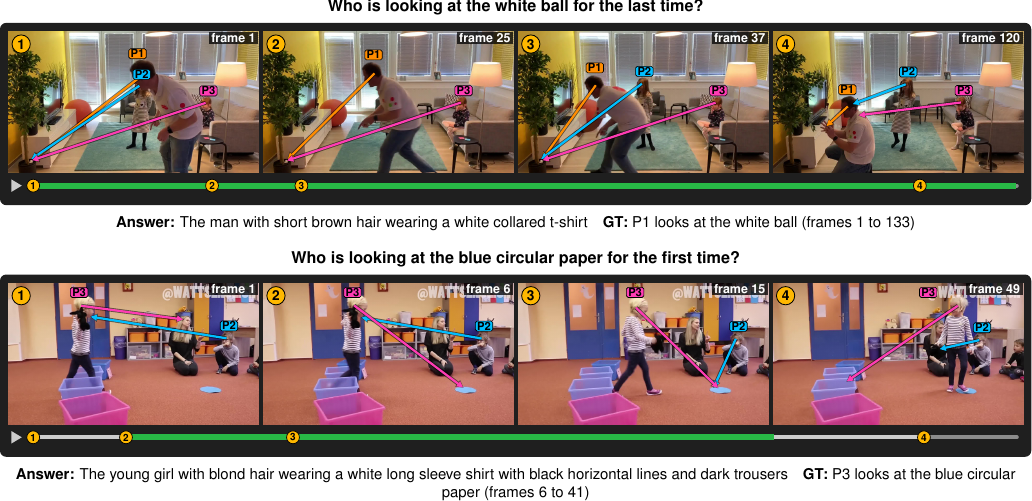}
\caption{\textbf{First or last person to look.} The question asks who looks at an object first or last. The answer names the people by appearance, and the GT gives the frames of their looks.}
\label{fig:supp-t03_first_last_looker}
\end{figure*}

\begin{figure*}[p]
\centering
\includegraphics[width=\textwidth]{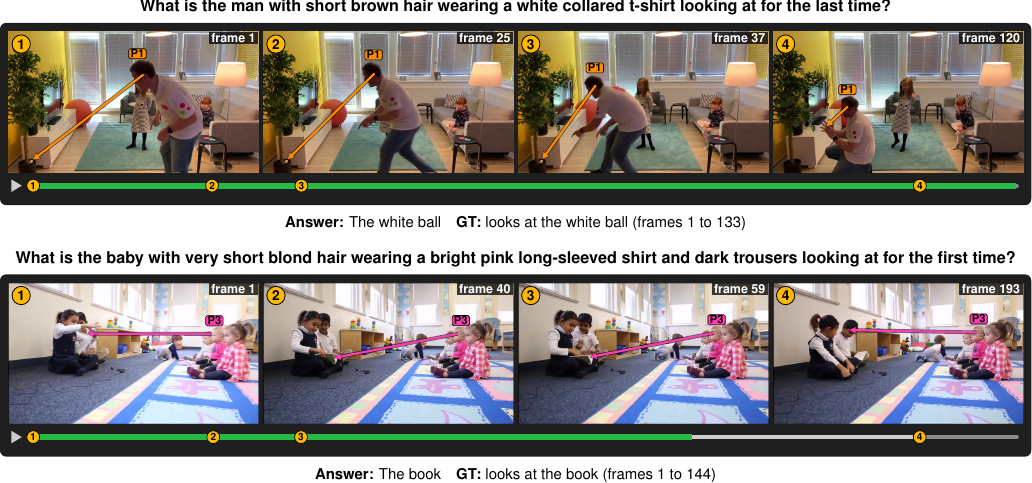}
\caption{\textbf{First or last target of a person.} The question asks what a person looks at first or last in the clip, so the VLLM has to find the beginning or the end of the gaze history of one person.}
\label{fig:supp-t04_person_first_last_target}
\vspace{14pt}
\includegraphics[width=\textwidth]{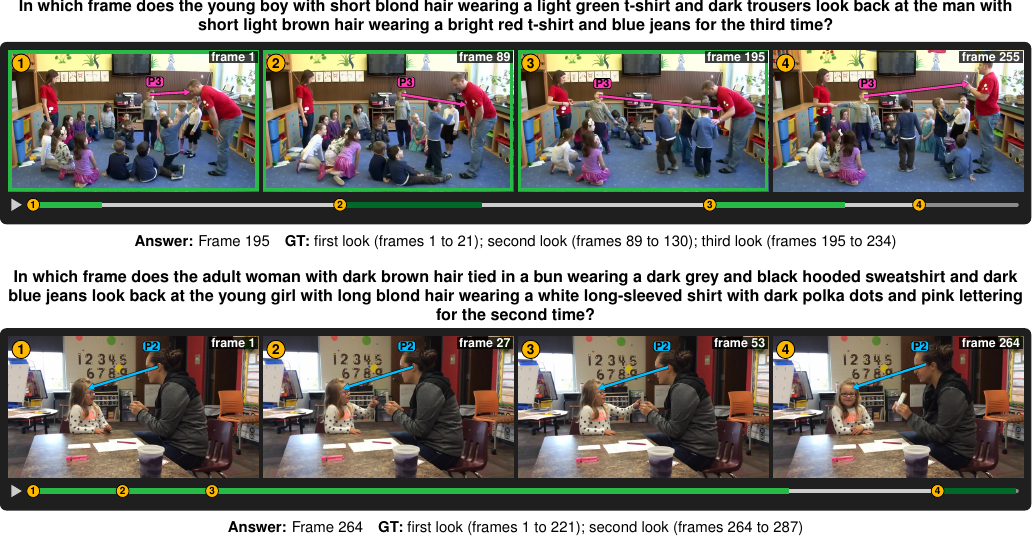}
\caption{\textbf{N-th look at an object.} The question asks in which frame a person looks back at an object for the second or third time. The GT lists every look in order, so the VLLM has to count separate looks and not frames.}
\label{fig:supp-t05_nth_look_at_object}
\end{figure*}

\begin{figure*}[p]
\centering
\includegraphics[width=\textwidth]{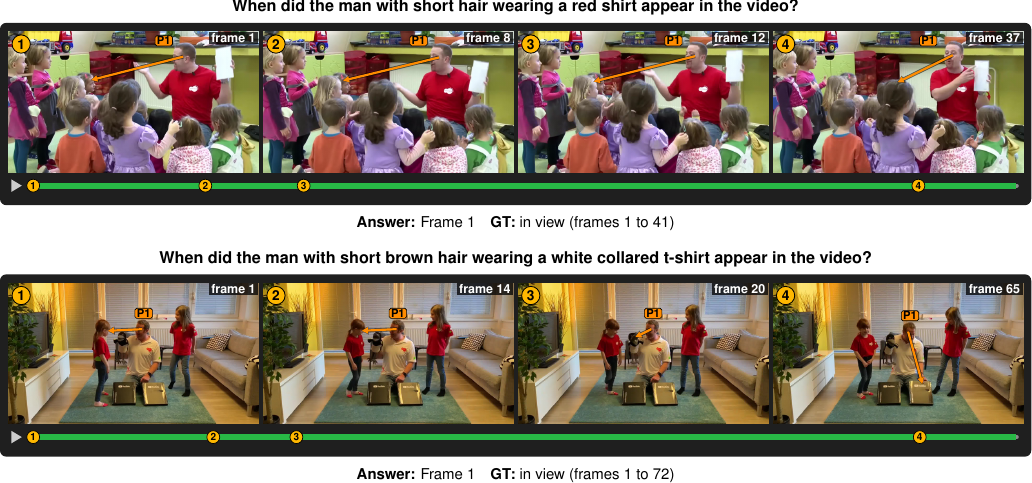}
\caption{\textbf{Appearance of a person.} The question asks in which frame a person first appears in the video. In both examples the person is in view from the start, so the answer is the first frame.}
\label{fig:supp-t06_person_appearance}
\vspace{14pt}
\includegraphics[width=\textwidth]{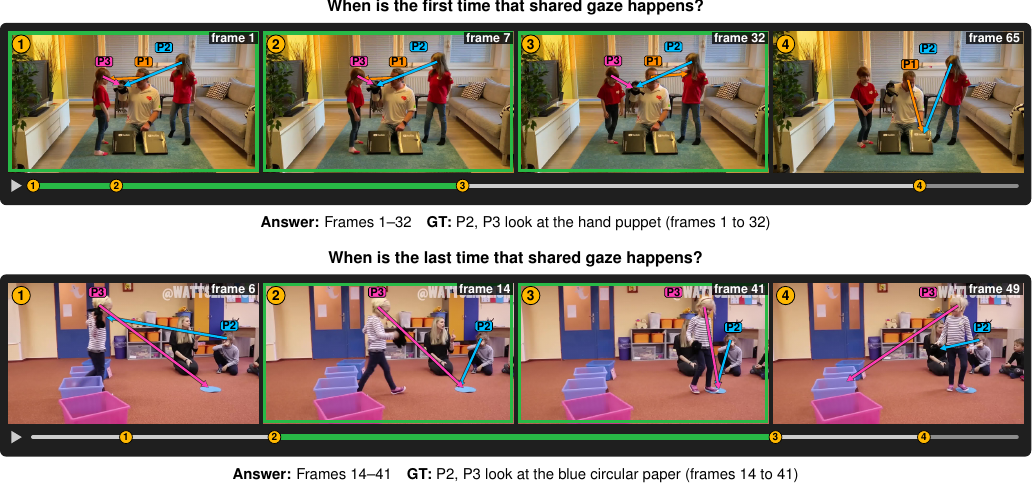}
\caption{\textbf{First or last shared gaze.} The question asks when several people look at the same target for the first or the last time. The GT names the people and the shared target.}
\label{fig:supp-t07_first_last_shared_gaze}
\end{figure*}

\begin{figure*}[p]
\centering
\includegraphics[width=\textwidth]{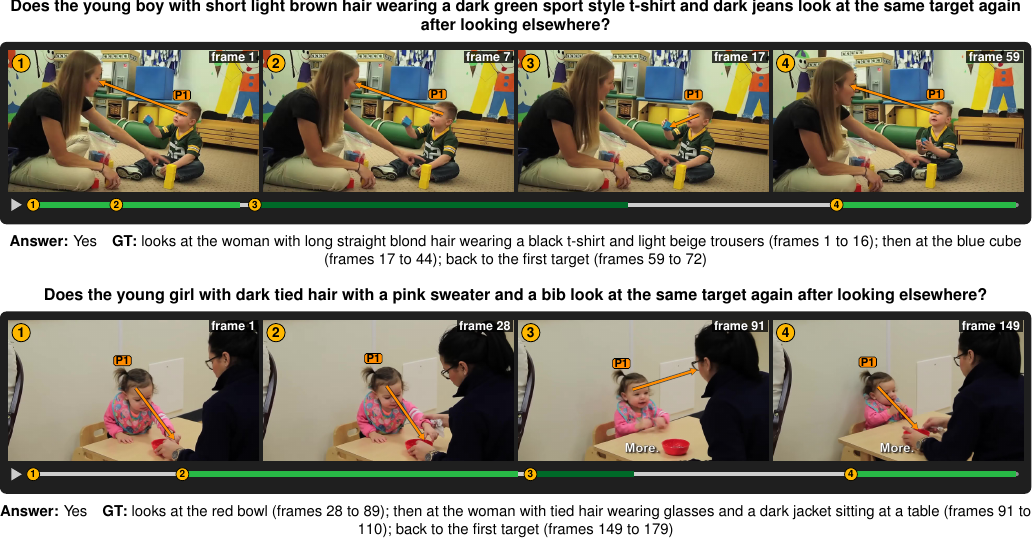}
\caption{\textbf{Return to a target.} The question asks whether a person looks at the same target again after looking elsewhere. The GT shows the first look, the look away and the return.}
\label{fig:supp-t08_gaze_target_revisiting}
\vspace{14pt}
\includegraphics[width=\textwidth]{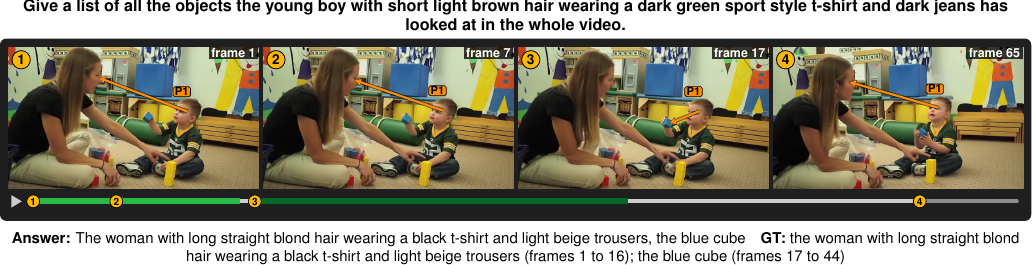}
\caption{\textbf{Objects a person looked at.} The question asks for every target of a person in the whole clip, and the answer lists them in the order of the looks.}
\label{fig:supp-t09_person_object_list}
\end{figure*}

\begin{figure*}[p]
\centering
\includegraphics[width=\textwidth]{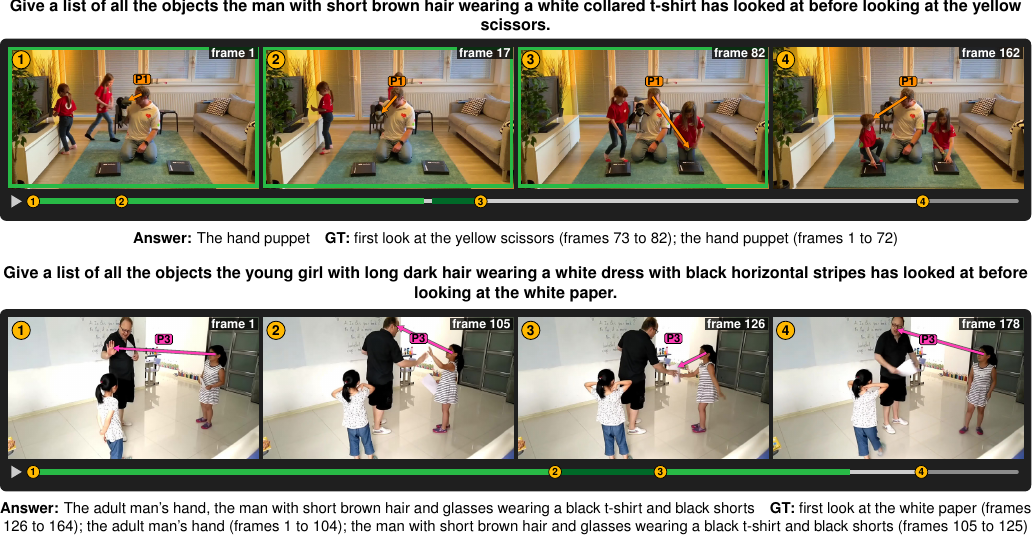}
\caption{\textbf{Objects before a target.} The question asks which targets a person looked at before a given object, so the VLLM has to keep the order of the looks. The GT starts with the first look at the given object.}
\label{fig:supp-t10_person_object_list_before}
\vspace{14pt}
\includegraphics[width=\textwidth]{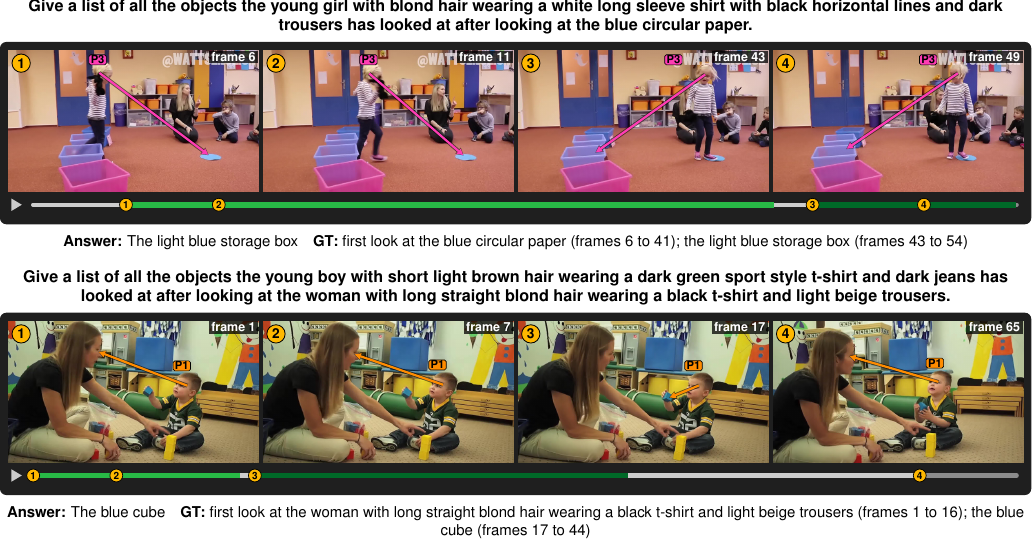}
\caption{\textbf{Objects after a target.} The question asks which targets a person looked at after a given object. The GT starts with the first look at the given object and lists the later looks.}
\label{fig:supp-t11_person_object_list_after}
\end{figure*}

\begin{figure*}[p]
\centering
\includegraphics[width=\textwidth]{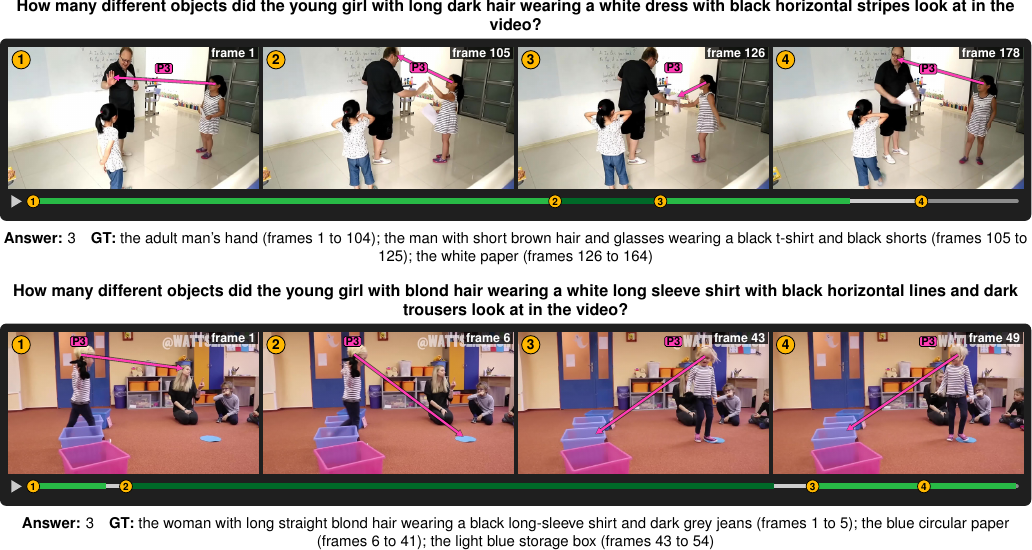}
\caption{\textbf{Number of targets of a person.} The question asks how many different objects a person looked at in the video, and the GT lists them with their frames.}
\label{fig:supp-t12_person_object_count}
\vspace{14pt}
\includegraphics[width=\textwidth]{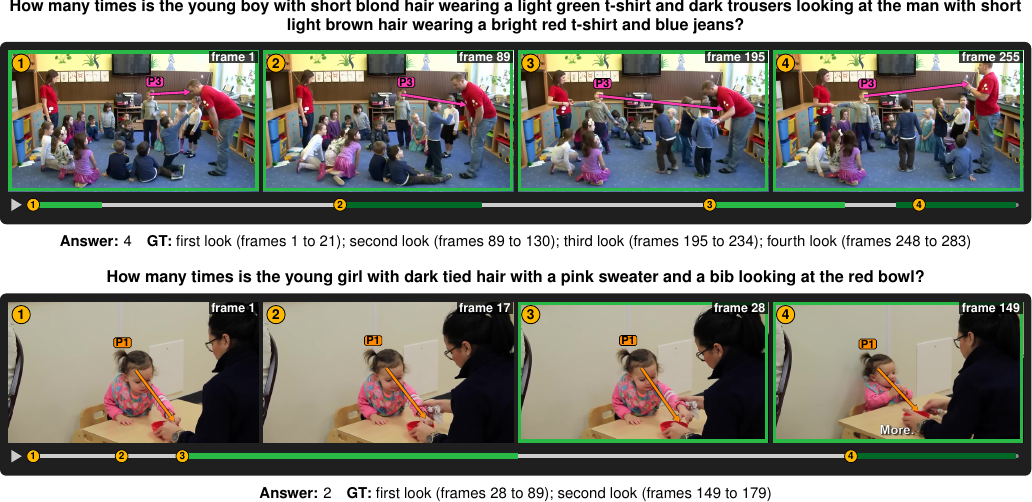}
\caption{\textbf{Number of looks of a person.} The question asks how many separate times a person looks at an object. A person who looks away and comes back starts a new look, so the answer counts looks and not frames.}
\label{fig:supp-t13_person_object_look_count}
\end{figure*}

\begin{figure*}[p]
\centering
\includegraphics[width=\textwidth]{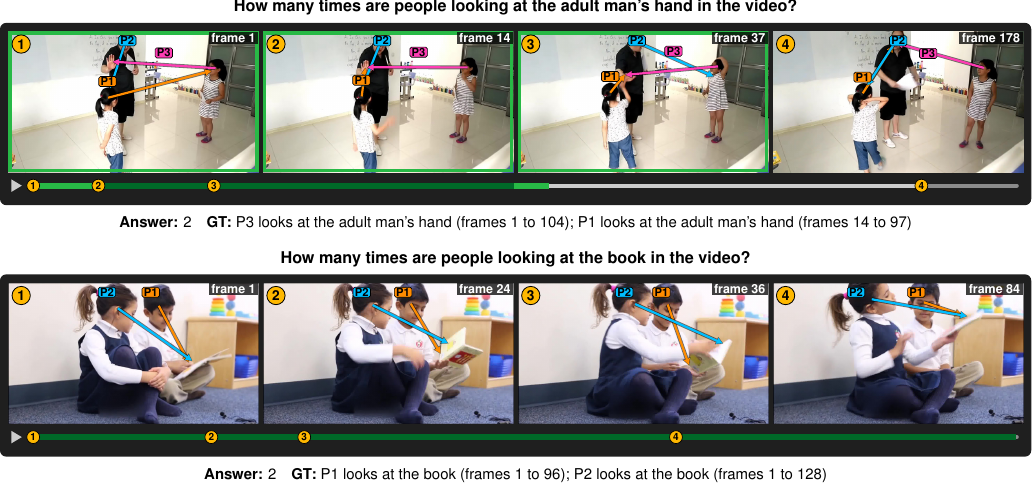}
\caption{\textbf{Number of looks at an object.} The question asks how many separate looks at an object happen in the video, counted over all people.}
\label{fig:supp-t14_object_look_count}
\vspace{14pt}
\includegraphics[width=\textwidth]{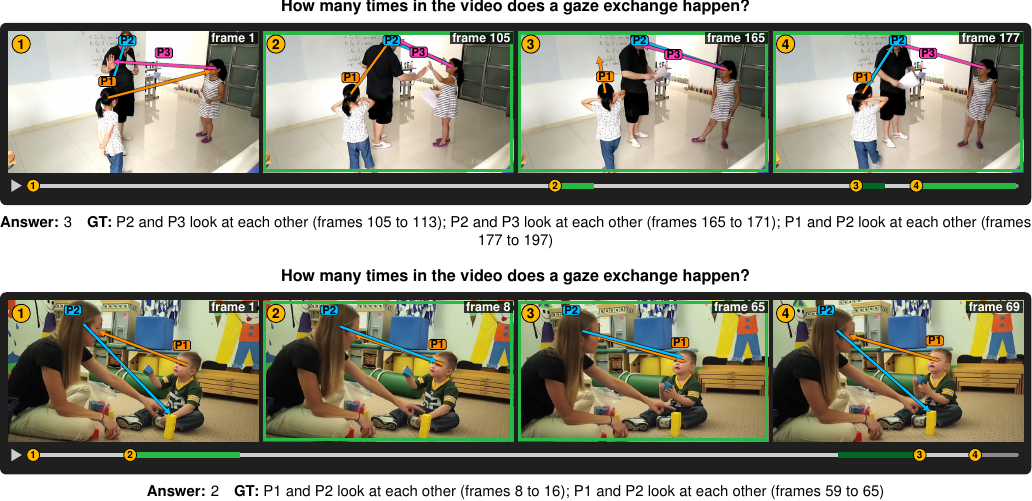}
\caption{\textbf{Number of gaze exchanges.} The question asks how many times two people look at each other in the video. The GT names the two people of every exchange.}
\label{fig:supp-t15_gaze_exchange_count}
\end{figure*}

\begin{figure*}[p]
\centering
\includegraphics[width=\textwidth]{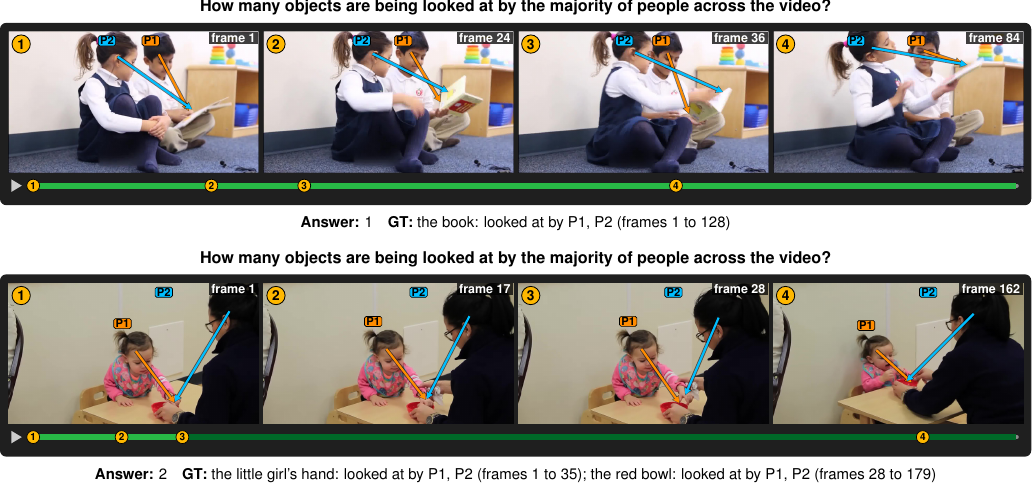}
\caption{\textbf{Objects looked at by most people.} The question asks how many objects are looked at by the majority of the people in the video. The GT names each object and the people who look at it.}
\label{fig:supp-t16_majority_object_count}
\vspace{14pt}
\includegraphics[width=\textwidth]{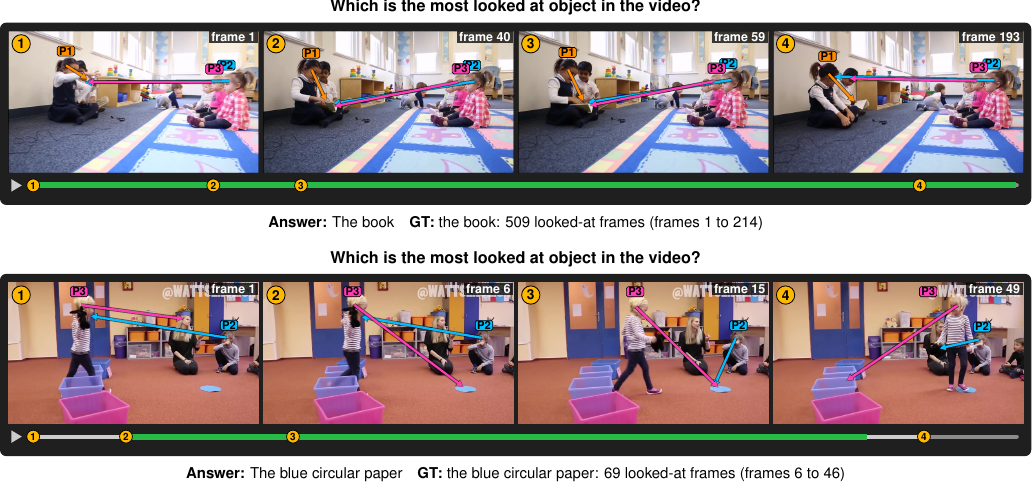}
\caption{\textbf{Most looked at object.} The question asks which object is looked at in the most frames. The GT gives this number summed over all people, so it can be larger than the length of the clip.}
\label{fig:supp-t17_most_looked_object}
\end{figure*}

\begin{figure*}[p]
\centering
\includegraphics[width=\textwidth]{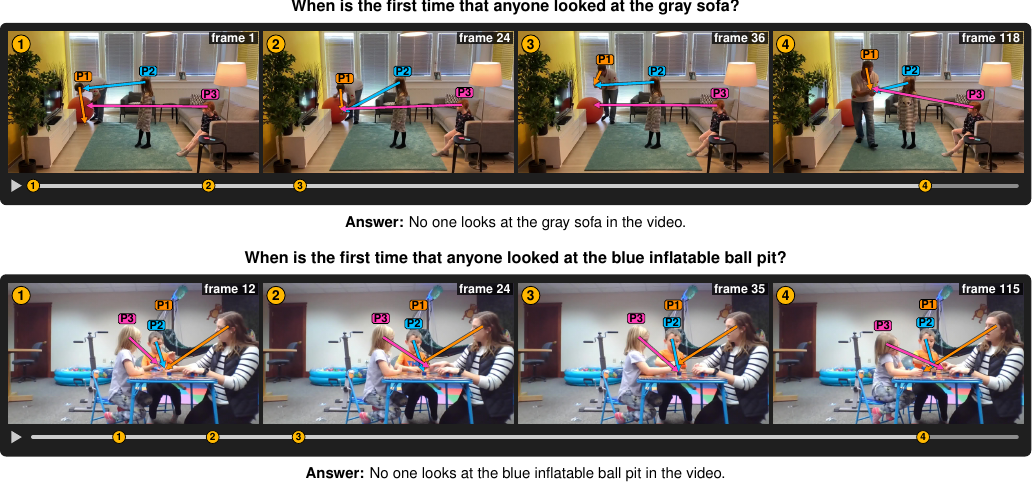}
\caption{\textbf{Object that nobody looks at.} The question asks when anyone first looked at an object that nobody looks at in the clip. The question has a false premise, and the correct answer says that nobody looks at the object.}
\label{fig:supp-t18_look_at_unlooked_object}
\vspace{14pt}
\includegraphics[width=\textwidth]{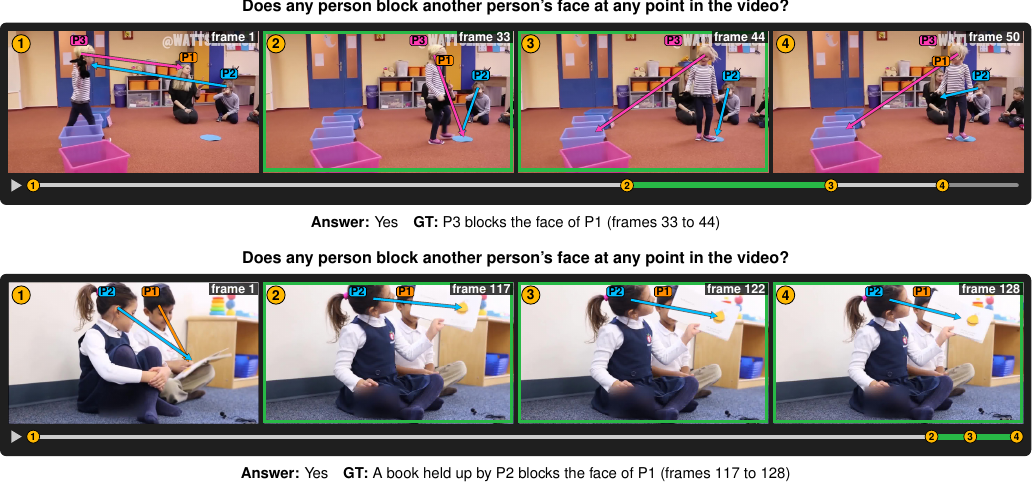}
\caption{\textbf{Blocked face.} The question asks whether one person blocks the face of another person at some point in the video. The GT is written from the annotation in plain words and keeps the same people and frames.}
\label{fig:supp-t19_face_occlusion_detection}
\end{figure*}

\begin{figure*}[p]
\centering
\includegraphics[width=\textwidth]{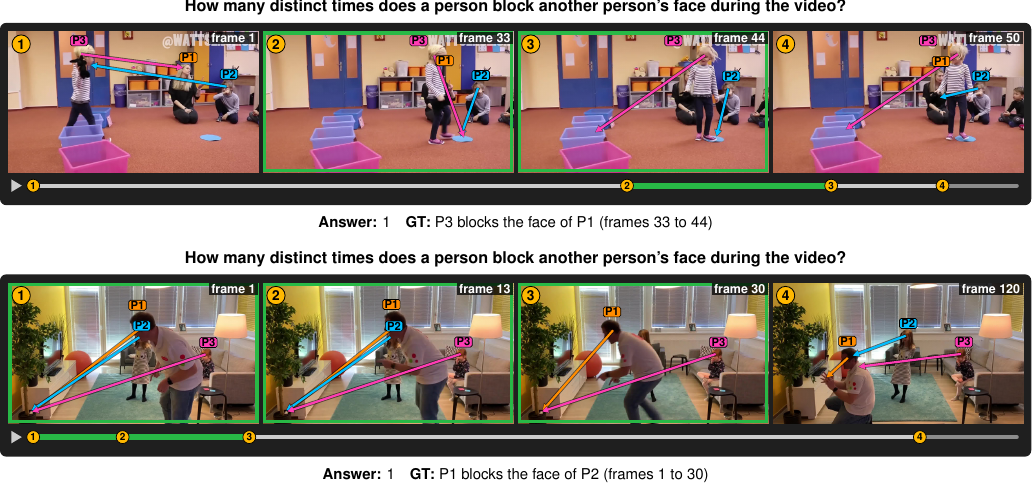}
\caption{\textbf{Number of blocked faces.} The question asks how many distinct times a person blocks the face of another person. Blocking events are found in the head boxes and confirmed by an annotator, who also marks who blocks whom and in which frames.}
\label{fig:supp-t20_face_occlusion_counting}
\vspace{14pt}
\includegraphics[width=\textwidth]{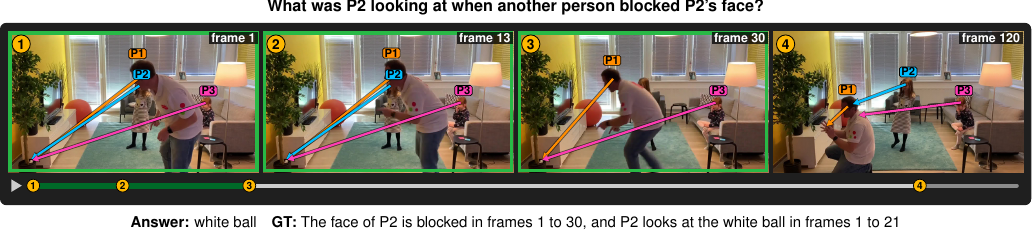}
\caption{\textbf{Target during a blocked face.} The question asks what a person looked at while another person blocked their face. The answer is the target that the annotators recorded for this person around the blocking event.}
\label{fig:supp-t21_occluded_gaze_reasoning}
\end{figure*}

\Cref{fig:supp-t01_first_frame_person_object} to \cref{fig:supp-t21_occluded_gaze_reasoning} show the 21 temporal types. They start with questions about when a look happens, continue with questions that list or count looks, targets, and gaze exchanges, and end with the questions about blocked faces.


\clearpage   
\section{Input Ablations and Per-Task Analysis}
\label{sec:supp-ablation}

In the main paper, each model sees an arrow from a person's head to the
gaze target predicted by our PaGE agent. If the model names the wrong
target, the answer alone does not tell us why. The arrow may point to the
wrong object, or the model may misread an accurate arrow. We use the same
questions with different inputs to study these two sources of error.

\paragraph{Inputs and scoring}
We test how much four models can tell about gaze from the question alone,
from the original frame or video, and from the same visual input with a gaze
arrow. We use two versions of the arrow. Both start at the person's head and end at either the gaze
point predicted by PaGE or the point marked by a human annotator.
We use the same 32{,}901 questions throughout. We average accuracy across
seven question types for frames and 21 for videos, giving each type equal
weight within its mean. PaGE scores come from the main-paper runs.
We measure the effect of adding an arrow by comparing
annotated arrows with unmarked inputs.

\subsection{What can models answer from the question alone?}
\label{sec:supp-ablation-inputs}

Gaze-target questions make up 73\% of the frame questions. Models answer
fewer than 1\% correctly from the question alone. Showing them the original
frame raises accuracy to 24--36\%, even before we add an arrow.
A baseline that predicts the most common answer for each question type in
the other clips also identifies fewer than 1\% of gaze targets. It answers
every absent-object question correctly because those questions share the same
answer. The need for visual information therefore varies by question type.

\subsection{Which questions benefit from a gaze arrow?}
\label{sec:supp-ablation-tasks}

All four models identify gaze targets more accurately when an arrow points
to the annotated location (Table~\ref{tab:supp-effects}). The gain is
clearest when the question asks what one person is looking at. The results
are more mixed when the answer depends on several people or targets.

\paragraph{Shared and exchanged gaze}
Two people can look at the same object (shared gaze) or at each other
(exchanged gaze). Both Qwen models improve on group questions overall.
For example, adding arrows raises Qwen3.5-2B accuracy by 28.8 percentage
points when it must identify people looking at each other.

\paragraph{Ambiguous questions}
Some questions have no unique target or ask for a count whose correct
answer is zero. On this group, three models lose accuracy with arrows.
InternVL3.5-14B is the only model that improves. An arrow is therefore
a more reliable cue when there is a specific target to identify.

\begin{table}[t]
\centering
\footnotesize
\captionsetup{font=footnotesize,width=\linewidth}
\caption{\textbf{Effect of annotated arrows on frame accuracy (\%).}
Red marks the highest value
per column in each block.}
\label{tab:supp-effects}
\setlength{\tabcolsep}{3pt}
\renewcommand{\arraystretch}{1.15}
\begin{tabular}{@{}p{\dimexpr0.32\linewidth-\tabcolsep\relax}*{3}{>{\centering\arraybackslash}p{\dimexpr0.2266\linewidth-2\tabcolsep\relax}}}
\toprule
\textbf{Model} & \textbf{Gaze target} & \textbf{Group} & \textbf{Ambiguous} \\
\specialrule{\lightrulewidth}{\aboverulesep}{0pt}
Qwen3.5-2B      & 26.3 & 37.1 & 44.9 \\
\hspace{1cm} +arrows & $+11.5$ & \cellcolor{rank1}$+7.9$ & $-8.9$ \\ \hline
Qwen3.5-35B-A3B & \cellcolor{rank1}35.8 & 42.8 & \cellcolor{rank1}73.5 \\
\hspace{1cm} +arrows & \cellcolor{rank1}$+15.5$ & $+5.8$ & $-11.4$ \\\hline
InternVL3.5-14B & 23.7 & \cellcolor{rank1}43.2 & 46.9 \\
\hspace{1cm} +arrows  & $+7.0$ & $+1.5$ & \cellcolor{rank1}$+3.4$ \\\hline
Molmo2-8B       & 24.5 & 28.4 & 18.4 \\
\hspace{1cm} +arrows & $+7.6$ & $-2.0$ & $-7.3$ \\\hline
\end{tabular}
\end{table}

\subsection{Do arrows help models follow gaze over time?}
\label{sec:supp-ablation-video}

\begin{figure*}[t]
\centering
\captionsetup{font=footnotesize,width=\linewidth}
\includegraphics[width=\linewidth]{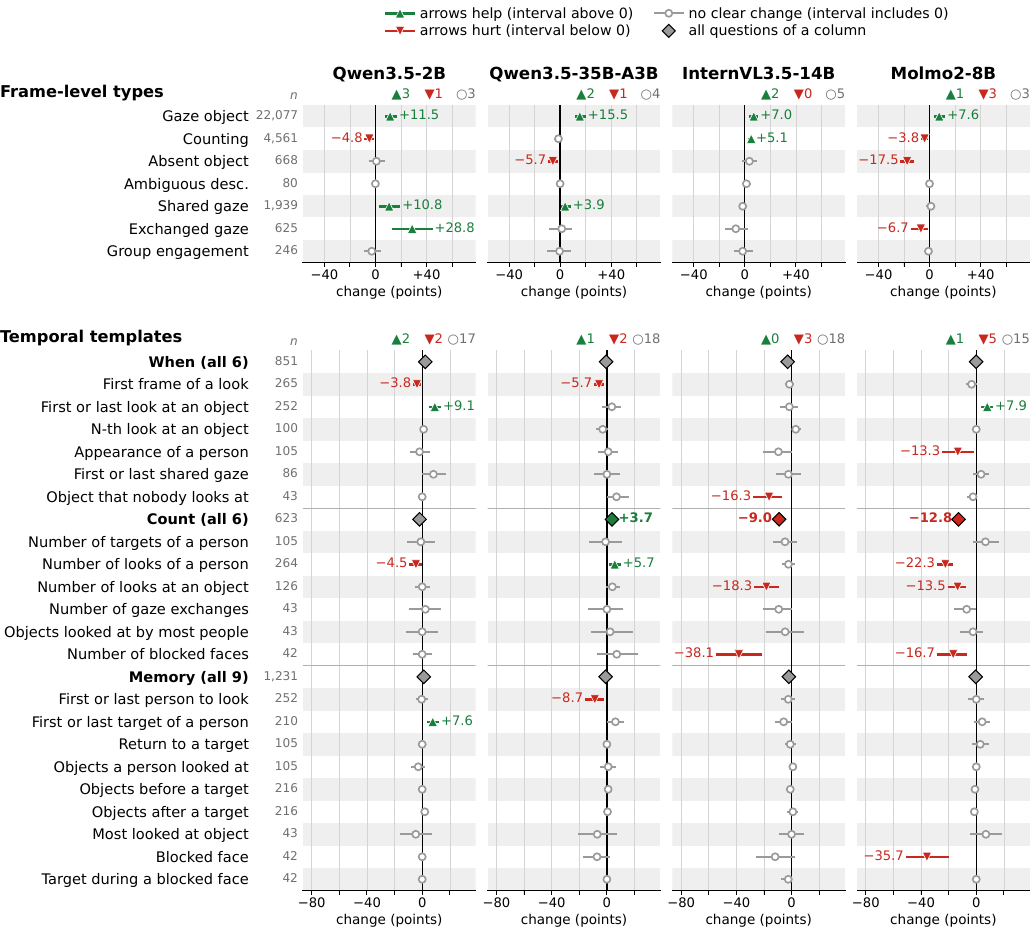}
\caption{\textbf{Arrow effects by question type.}
Each row gives the change in accuracy from annotated arrows over no arrows
in the same runs, in percentage points. Bars show paired 95\% confidence
intervals from 10{,}000 resamples of the 43 test clips. Green and red mark
intervals above and below zero, and grey circles mark intervals that include
zero. Diamonds summarize the When, Count, and Memory columns of the main
paper, and grey numbers give the questions per row.}
\label{fig:supp-effects}
\end{figure*}

An arrow marks where a person is looking in one frame. To answer when
someone first looks at an object or how often they return to it, a model
must follow that person's gaze across frames. Depending on the model, 15 to 18 of the 21 video question types show
no clear change in accuracy with arrows (Figure~\ref{fig:supp-effects}).
The benefit of marking a gaze target is clearest in individual frames.

\subsection{How much does the arrow's position matter?}
\label{sec:supp-ablation-qwen}

We group gaze-target questions by how far the PaGE arrow tip falls from
the annotated point. Within each group, we compare the annotated arrow
with the same frame shown without an arrow. All four models gain more
when PaGE misses by over 100 pixels than when it lands within 50 pixels
(Table~\ref{tab:supp-distance-gains}). The largest difference is for
Qwen3.5-35B-A3B, whose gain rises from 10.9 to 25.0 percentage points.

\begin{figure*}[t]
\centering
\captionsetup{font=footnotesize,width=\linewidth}
\includegraphics[width=\linewidth]{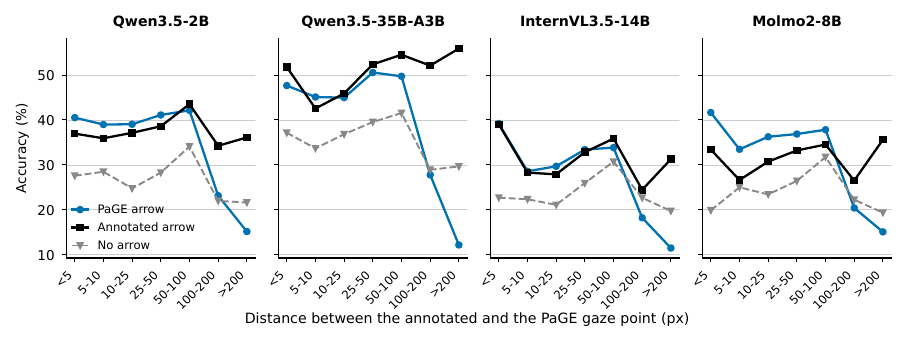}
\caption{\textbf{Gaze-target accuracy by tracking error.}
Distance is measured from the PaGE arrow tip to the annotated gaze point.}
\label{fig:supp-distance}
\end{figure*}

\begin{figure*}[t]
\centering
\captionsetup{font=footnotesize,width=\linewidth}
\includegraphics[width=\linewidth]{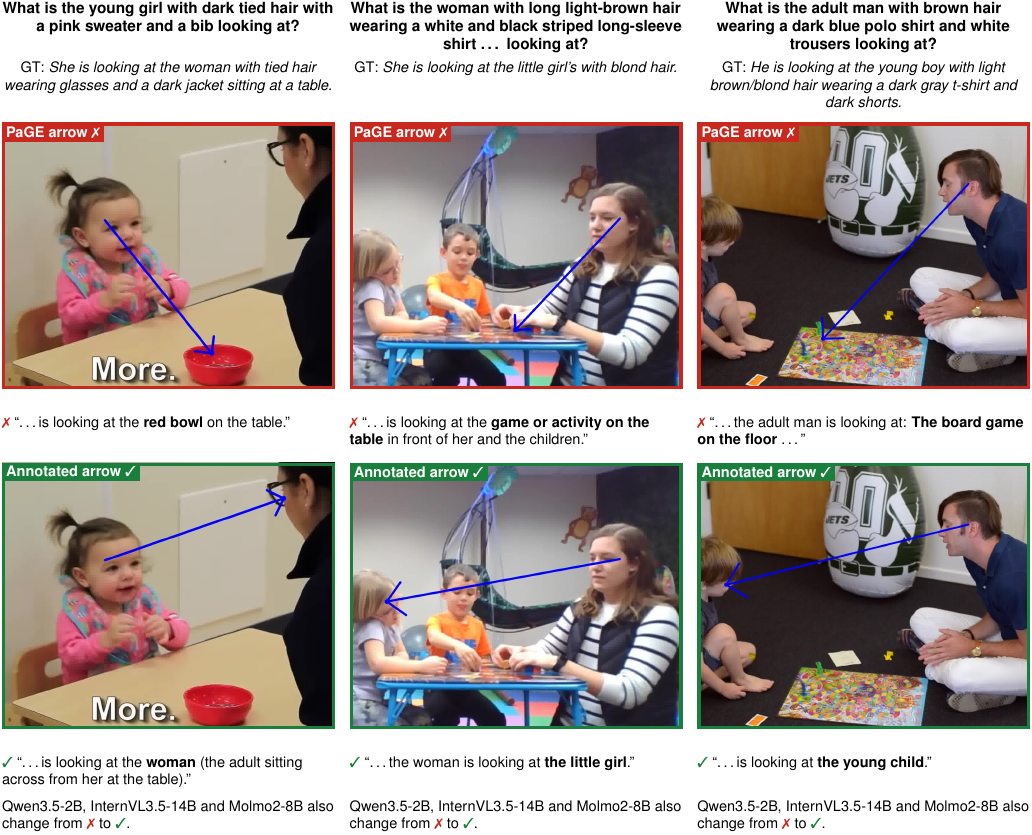}
\caption{\textbf{How answers vary with arrow position.}
Qwen3.5-35B-A3B answers the same questions with PaGE arrows (top, red)
and annotated arrows (bottom, green). In each example the PaGE arrow ends
far from what the person looks at. The images are crops of the model inputs.}
\label{fig:supp-gt-examples}
\end{figure*}

On the larger misses, all four models score lower with PaGE arrows in
the main-paper runs than with unmarked frames in the later runs
(Figure~\ref{fig:supp-distance}). In Figure~\ref{fig:supp-gt-examples},
Qwen3.5-35B-A3B names the board game when the
PaGE arrow points to it and the boy when the annotated arrow points to
him. In this example, the model follows the arrow in both cases, so where
the tracker places its tip matters for the answer.

\begin{table}[h]
\centering
\footnotesize
\captionsetup{font=footnotesize,width=\linewidth}
\caption{\textbf{Gaze-target gains by PaGE tracking error.}
Values show gains from annotated arrows over no arrows (pp).
Red marks the highest gain per column.}
\label{tab:supp-distance-gains}
\setlength{\tabcolsep}{3pt}
\renewcommand{\arraystretch}{1.15}
\begin{tabular}{@{}p{\dimexpr0.32\linewidth-\tabcolsep\relax}*{2}{>{\centering\arraybackslash}p{\dimexpr0.34\linewidth-2\tabcolsep\relax}}}
\toprule
\textbf{Model} & \textbf{At most 50 px} & \textbf{Over 100 px} \\
\midrule
Qwen3.5-2B      & \cellcolor{rank1}$+10.9$ & $+13.6$ \\
Qwen3.5-35B-A3B & \cellcolor{rank1}$+10.9$ & \cellcolor{rank1}$+25.0$ \\
InternVL3.5-14B & $+7.2$  & $+7.5$ \\
Molmo2-8B       & $+6.9$  & $+11.3$ \\
\bottomrule
\end{tabular}
\end{table}

\end{document}

%% file: preamble.tex
\renewcommand{\paragraph}[1]{\vspace{.5em}\noindent\textbf{#1.}}

\usepackage{xspace}

\newcommand{\wgaze}{\texorpdfstring{W\textsuperscript{3}Gaze~VQA}{W3Gaze VQA}\xspace}

\usepackage{colortbl}   
\usepackage{cuted}      
\definecolor{rank1}{HTML}{FF9E9E}  
\definecolor{rank2}{HTML}{FFC78A}  
\definecolor{rank3}{HTML}{FFF08A}  

\newcommand{\yes}{\cellcolor{green!25}\rule[-0.7ex]{0pt}{3.0ex}\ensuremath{\checkmark}}
\newcommand{\no}{\cellcolor{red!25}\rule[-0.7ex]{0pt}{3.0ex}\ensuremath{\times}}

\usepackage{soul}
\setuldepth{foobar}

%% file: sec/0_abstract.tex
\begin{abstract}

Social gaze understanding goes beyond the prediction of gaze direction. It requires knowing who is looking at what and how shared attention changes among people over time. Vision Large Language Models (VLLMs) can answer such questions. However, current gaze benchmarks do not test their capability since they mainly concentrate on gaze localisation in static images or use automatically annotated datasets, while overlooking the temporal nature of attention changes. In this work, we introduce \wgaze, the first manually annotated video benchmark for social and temporal gaze reasoning, asking \textbf{w}ho looks at \textbf{w}hat and \textbf{w}hen. The dataset consists of 381 video clips, with around 7K manually annotated key frames that yield around 640K frame-level QA pairs across nine question types with around 16K video-level temporal questions. All appearance, gaze, and social attention annotations were created and verified by human annotators. We also provide ground truth corrections for the gaze annotations of 50 ChildPlay videos. Finally, we introduce a training-free gaze tracking agent that needs 85\% fewer estimator calls than estimating gaze in every frame. Across 16 VLLMs, the strongest model answers only 41\% of the questions correctly, and average accuracy falls from 32\% on frame-level questions to 21\% on temporal questions, showing that following gaze attention through time remains an open challenge. The code is publicly available \href{https://github.com/shayan137/GazeVQA}{here}.

\end{abstract}

%% file: sec/1_intro.tex
\section{Introduction}

Understanding human gaze is fundamental to interpreting social scenes, because gaze shows what people attend to and how they direct and respond to attention towards objects and others~\cite{HENDERSON2003498, okumura2013power}. Considerable research attention has been devoted to gaze following and gaze target estimation, which infer where a person is attending within a visual scene~\cite{gazefollow, 10.1007/978-3-030-01228-1_24}. Most of these methods treat gaze as a localisation problem and predict a point for an individual in static images~\cite{ryan2025gazellegazetargetestimation, Tafasca_2024_CVPR, JIN2022104924}, and newer benchmarks extend this to scenes with several people~\cite{chong2020detectingattendedvisualtargets, Tafasca_2023_ICCV}. On the other hand, vision large language models (VLLMs) present new opportunities to go further from gaze localisation into a richer understanding of social attention~\cite{nasiriboukani2025gaze, TowardSemantic}, enabling them to address questions that gaze points alone cannot answer, for instance, who a person attends to, whether several people share a target, or whether someone looks back at an object they saw earlier. However, assessing the ability of VLLMs to answer such questions in real-life social scenes is challenging for three main reasons.

\paragraph{First, social gaze reasoning has no reliable benchmark} While predicting a point indicates where a person is looking, it does not tell us what a person is looking at, if two people are looking at each other, or if a group is jointly engaged. GazeAnywhere~\cite{Cao_2026_CVPR} identifies a person based on their appearance and predicts where they are looking, but still provides only a gaze location. Recent work on gaze visual question answering extends gaze estimation beyond spatial locations by identifying the objects people are looking at. However, existing approaches are limited to static images and rely heavily on automatically generated questions and answers, where generation errors can affect both the supervision and evaluation of VLLMs~\cite{wang2025vl4gazeunleashingvisionlanguagemodels}. 

\paragraph{Second, gaze changes over time, which single-frame questions cannot capture} These social questions become harder in video, where people move, appearances change, and a VLLM must also tell whether and how often one person blocks another person's face. Answering when a person first looked at an object, how many times they looked at it, or who looked at it last requires remembering what happened in earlier frames and ordering these looks across the clip. Existing video models often lose track of objects when they disappear and reappear~\cite{chen2026toc}, struggle to distinguish between similar-looking objects~\cite{liu2026can}, and perform much worse than humans at remembering events and their order~\cite{cai2024temporalbench, yang2025thinking}. No gaze benchmark tests these abilities, because image-based resources cannot ask such questions and video resources such as ChildPlay were not designed for natural language reasoning and lack appearance and social relation annotations~\cite{Tafasca_2023_ICCV}.

\paragraph{Third, video gaze is costly to estimate} Both types of questions depend on knowing where each person is looking, but existing gaze estimators work on individual images~\cite{ryan2025gazellegazetargetestimation, ye2026page}. Running them for every person in every video frame takes time and repeats the same work when a person's gaze remains on the same target.

To address these challenges, we introduce \wgaze, a video benchmark for social and temporal gaze reasoning that asks \textbf{w}ho looks at \textbf{w}hat, and \textbf{w}hen. We build the benchmark using naturalistic, child-centred ChildPlay videos (\cref{fig:gaze_vqa_examples}) and develop a training-free agent to track gaze across frames. Our evaluation of 16 leading VLLMs shows that even the best model answers only 41\% of the questions correctly, with average accuracy dropping from 32\% on frame-level questions to just 21\% on temporal questions (\cref{tab:w3_gaze_VQA_results}).

Our \textbf{contributions} are threefold:
\begin{itemize}

    \item \textbf{Social gaze reasoning:} We build \wgaze with around 640K human verified question-answer pairs over 381 clips in nine question types, including counting, absent object, and ambiguous description questions, and we correct the gaze labels of 50 ChildPlay videos.
    
    \item \textbf{Temporal gaze reasoning:} We introduce around 16K questions across 21 different types that explore how people's gaze changes throughout a video, from when and how often they look at something to whether they return to it later or block another person's face.
    
    \item \textbf{Efficient gaze prompting:} We develop a training-free agent that estimates each person's gaze target and tracks it across video frames, updating the prediction only when a face analysis model detects a significant change in head or eye behaviour~\cite{hu2025openface30lightweightmultitask}. This cuts estimator calls by 85\% while providing visual gaze prompts to VLLMs.
    
\end{itemize}

%% file: sec/2_related_work.tex
\section{Related Work}

\subsection{Human Gaze Target Estimation} Understanding where people direct their attention is a core problem in gaze research, and early gaze estimation methods addressed it using scene priors and activity-specific knowledge to narrow the set of possible gaze targets, but these assumptions limited their ability to generalise to unconstrained environments~\cite{gazefollow}. The field therefore shifted towards directly predicting gaze locations from visual inputs, leading to the widely adopted multi-branch paradigm in which the target person and the surrounding scene are processed separately and later fused~\cite{10.1007/978-3-030-01228-1_24,chong2020detectingattendedvisualtargets,9666980,8593580,9828503}. As these models matured, researchers began incorporating additional cues to resolve ambiguities that could not be handled from RGB appearance alone. Depth-aware methods introduced geometric information to distinguish objects that appear close in the image but lie at different distances in the scene~\cite{JIN2022104924,10.1145/3536221.3556624,miao2022patchlevelgazedistributionprediction,Horanyi_2023_CVPR}. Pose-aware approaches then incorporated body orientation and skeletal information, providing additional evidence about a person's likely direction of attention when head appearance alone was ambiguous~\cite{6027285,10.1007/978-3-031-91575-8_22,gupta2023modularmultimodalarchitecturegaze}. This was further extended by 3D head and gaze modelling, where head orientation and gaze direction were represented in three-dimensional space to better constrain potential targets in cluttered or spatially complex scenes~\cite{9879695,Fang_2021_CVPR}. Other works focused more closely on facial and eye information, combining the face and individual eye regions to improve gaze direction estimation under challenging head poses, occlusions, and limited eye visibility~\cite{10.1093/jcde/qwaf017,cheng2020coarsetofineadaptivenetworkappearancebased,9050633,cai2021gazeestimationensemblearchitectures,chen2019appearancebasedgazeestimationusing,10.1093/jcde/qwad038,5959981,10.1007/978-3-030-01249-6_21}. More recent work has broadened gaze following through object-aware and end-to-end Transformer formulations that jointly reason about people, scene context, and potential gaze targets~\cite{Tonini_2023_ICCV,Tu_2022_CVPR,Tafasca_2024_CVPR}. Temporal and multi-person approaches have further extended gaze estimation towards dynamic scenes and interactions among multiple individuals~\cite{gupta2024mtgs}.

\begin{figure*}[t]
    \centering
    \includegraphics[width=\textwidth]{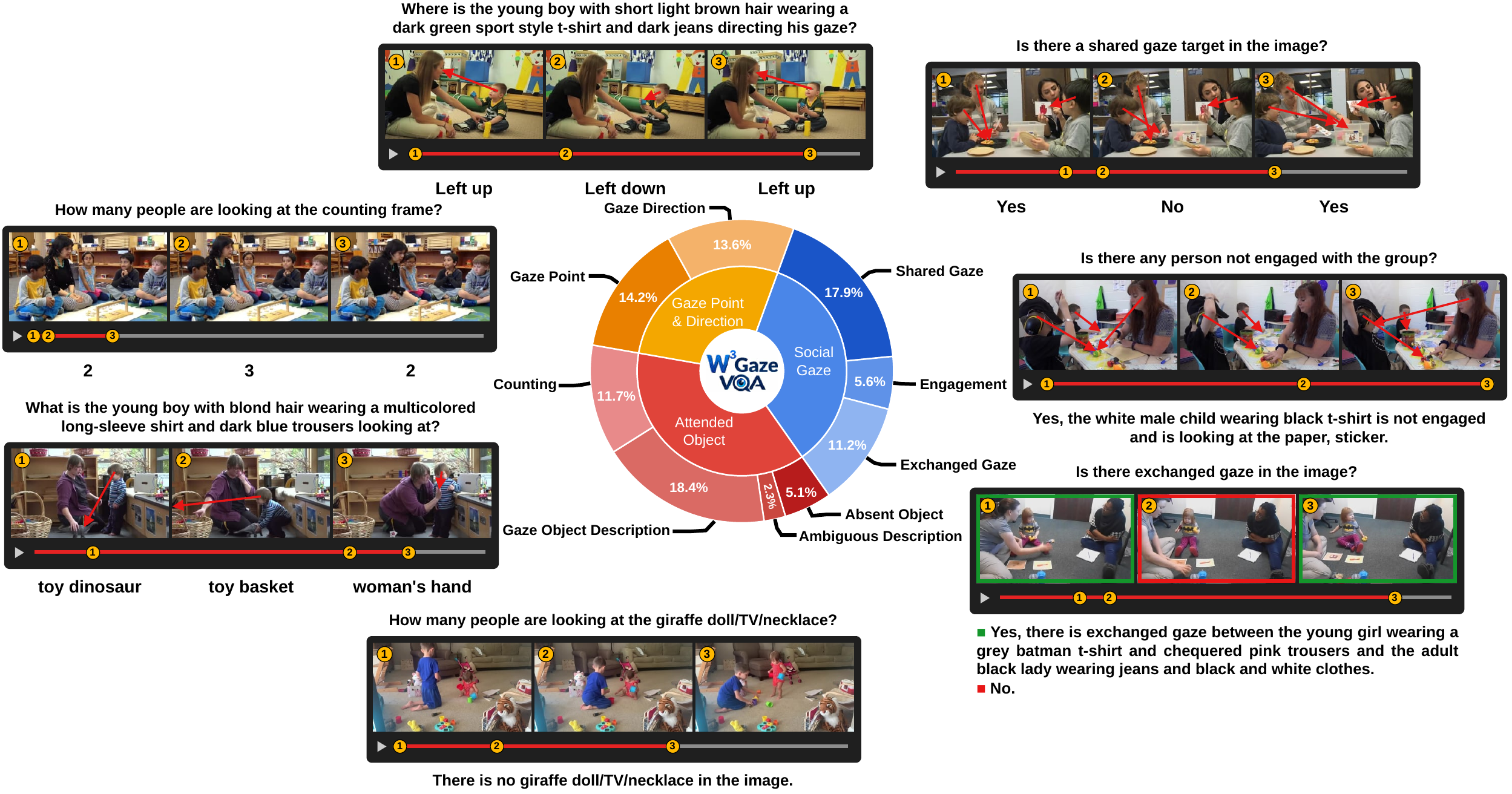}
    \caption{
    \textbf{Overview of the \wgaze benchmark tasks and dataset composition.}
    The benchmark has three main categories (gaze point and direction, 
    attended object, and social gaze) with nine question types. Representative examples illustrate 
    gaze direction, counting, gaze object description, 
    absent object, shared gaze, exchanged gaze, and engagement questions.
    }
    \label{fig:g3vqa_tasks}
\end{figure*}

\subsection{Multimodal Foundation Models for Gaze Understanding} Vision and vision-language foundation models have enabled strong transferable, open-vocabulary, and zero-shot visual understanding~\cite{radford2021learning,li2023blip,NEURIPS2023_6dcf277e,bai2025qwen3}. This progress has also begun to reshape gaze understanding, where pretrained visual encoders provide stronger person and scene representations and semantic approaches move beyond gaze localisation towards identifying what is being attended to~\cite{tu2023jointgazelocationgazeobjectdetection,TowardSemantic}. GazeLLM and GGVL further explore language-based reasoning for gaze, using LLMs or VLLMs to infer attended targets and support richer semantic and social understanding~\cite{gazellm,nasiriboukani2025gaze}. In parallel, GazeAnywhere introduces a promptable formulation in which the target person can be specified through semantic concepts rather than only predefined spatial inputs~\cite{Cao_2026_CVPR}. Despite these advances, existing methods still focus mainly on locating or identifying gaze targets, leaving broader multimodal reasoning about gaze, people, and their interactions relatively underexplored.

\subsection{VQA for Gaze Understanding} VQA provides a flexible way to test whether VLLMs truly understand a visual scene through targeted questions~\cite{Ganz_2024_CVPR}. This is important because strong overall performance can hide basic failures such as hallucinating objects, relying on language priors, or answering incorrectly despite clear visual evidence~\cite{Guan_2024_CVPR,Tong_2024_CVPR,Liu_2025_CVPR}. VLLMs also continue to struggle with counting, spatial relations, and complex multi-object scenes~\cite{Jain_2024_CVPR,Wang_2025_CVPR}, as well as multi-person and social reasoning~\cite{Lee_2024_CVPR}.
Video makes this harder, as VLLMs lose an object that is hidden and comes back~\cite{chen2026toc}, drop to chance when they must follow one of several identical objects~\cite{liu2026can}, and fall far below humans at event order and memory within a clip~\cite{cai2024temporalbench,yang2025thinking}.

\subsection{Object Permanence and Predictive Gaze} These failures across frames point to a missing world model: an internal picture of the scene that persists when something is out of sight and predicts what comes next~\cite{ha2018world,lecun2022path,hafner2023mastering}. Video models now learn such a picture at scale, as predicted features~\cite{assran2025v} or generated video~\cite{agarwal2025cosmos}. The classic test of whether a model keeps such a picture is object permanence, the expectation that a hidden object is still there, which infants show from a few months of age~\cite{piaget2013construction,baillargeon1985object}. Benchmarks test the same expectation through intuitive physics~\cite{riochet2018intphys} and tracking through occlusion~\cite{girdhar2019cater,shamsian2020learning}, and models learn it from video by design~\cite{piloto2022intuitive} or on their own~\cite{garrido2025intuitive}. World model benchmarks for VLLMs ask for memory and prediction~\cite{he2024mmworld}, but never about where people look. Gaze is a natural probe because it is predictive. The eyes reach an object before the hands do~\cite{land2001ways}, and they do so even when we watch someone else act~\cite{flanagan2003action}. Yet predicting future gaze has been studied only from the first person view~\cite{zhang2017deep}, third person work is just starting~\cite{kuang2026sage}, and theory of mind benchmarks track beliefs over time only in synthetic homes~\cite{jin2024mmtom,shi2025muma}. Our \wgaze benchmark evaluates temporal and occluded target reasoning, requiring memory of past gaze targets and of their order, and counting separate looks rather than frames.

%% file: sec/3_benchmark.tex
\section{\wgaze Benchmark}

Existing gaze datasets cover different parts of gaze understanding. Some focus on predicting where a person is looking in an image or video~\cite{gazefollow,recasens2017following,chong2020detectingattendedvisualtargets,Tafasca_2023_ICCV}, while others estimate gaze direction~\cite{Kellnhofer_2019_ICCV,Zhang_2020_ECCV}, identify the object being looked at~\cite{Tomas_2021_CVPR,TowardSemantic}, or allow users to select a person based on their appearance~\cite{Cao_2026_CVPR}. Other datasets focus on social gaze, such as whether people are looking at each other~\cite{marinjimenez2019laeonetrevisitingpeoplelooking,chong2020detectingattendedvisualtargets,gupta2024mtgs}. However, these datasets mainly provide gaze locations, directions, or object labels rather than questions about how people and objects are connected through gaze. Existing gaze question answering work is also limited to static images~\cite{wang2025vl4gazeunleashingvisionlanguagemodels}, leaving gaze reasoning across video frames largely unexplored. To address this gap, \wgaze introduces questions about individual people, interactions between people, and gaze changes throughout a video. It is also the first manually annotated\footnote{%
The source videos originate from ChildPlay~\cite{Tafasca_2023_ICCV},
whose collection and annotation were approved by the
original authors' Data and Research Ethics Committee.
According to national law, downloading YouTube videos
for training and annotation is permitted for research purposes.
For distribution, similarly to other datasets~\cite{gu2018ava},
we will share the video links, frame-extraction scripts,
and corresponding annotations rather than the original videos.} video benchmark for social and temporal gaze reasoning. \Cref{tab:gaze_dataset_comparison} compares \wgaze with existing gaze benchmarks.

\begin{table}[h]
\centering
\caption{\textbf{Comparison of \wgaze with existing gaze benchmarks.}
\wgaze combines gaze localisation, object and appearance understanding, gaze VQA, and social gaze VQA.}
\scriptsize

\renewcommand{\arraystretch}{1}
\setlength{\extrarowheight}{2pt}
\setlength{\tabcolsep}{2pt}

\makeatletter
\def\cellfill#1{\gdef\CT@cell@color{\CT@color{#1}\@tempdimb\dimexpr\tabcolsep+0.4pt\relax\@tempdimc\@tempdimb\global\let\CT@cell@color\relax}}
\makeatother
\def\yes{\cellfill{green!25}\rule[-0.7ex]{0pt}{3.0ex}\ensuremath{\checkmark}}
\def\no{\cellfill{red!25}\rule[-0.7ex]{0pt}{3.0ex}\ensuremath{\times}}
\def\hc#1{\setlength{\extrarowheight}{0pt}\renewcommand{\arraystretch}{1.15}\begin{tabular}[c]{@{}>{\bfseries}c@{}}#1\end{tabular}}

\resizebox{\columnwidth}{!}{%
\begin{tabular}{lcccccc}
\toprule
&
\hc{GazeFollow\\GazeHOI} &
\hc{ChildPlay\\VAT} &
\hc{Gaze-Co} &
\hc{GazeVQA} &
\hc{VL4Gaze} &
\hc{\wgaze} \\
\midrule
\noalign{\vskip-\belowrulesep} 

Gaze Coordinates
& \yes & \yes & \yes & \yes & \yes & \yes \\[-0.4pt]

Object Labels
& \yes & \no & \no & \yes & \yes & \yes \\[-0.4pt]

Person Appearance
& \no & \no & \yes & \no & \yes & \yes \\[-0.4pt]

Social Gaze
& \no & \yes & \no & \no & \no & \yes \\[-0.4pt]

Gaze VQA
& \no & \no & \no & \yes & \yes & \yes \\[-0.4pt]

\textbf{Social Gaze VQA}
& \no & \no & \no & \no & \no & \yes \\
\noalign{\vskip-\aboverulesep} 

\bottomrule
\end{tabular}%
}

\label{tab:gaze_dataset_comparison}

\end{table}

\subsection{Frame-Level Questions}
Frame-level questions refer to a single frame and are organised into nine types in three groups, with examples in \cref{fig:g3vqa_tasks} and Supp.\ \cref{sec:supp-frame-examples}. Each type posed to the VLLMs probes one specific weakness of current VLLMs, or one ability that we want them to acquire. 

\paragraph{Gaze point and direction} The first group contains gaze point and gaze direction questions, which ask where a person looks and in which direction. These two types are not posed to the VLLMs. We measure them with the gaze estimators and our tracking agent (\cref{tab:gaze_tracking_comparison}), which matters because the arrows shown to the VLLM are derived from them.

\begin{figure*}[t]
    \centering
    \includegraphics[width=\textwidth]{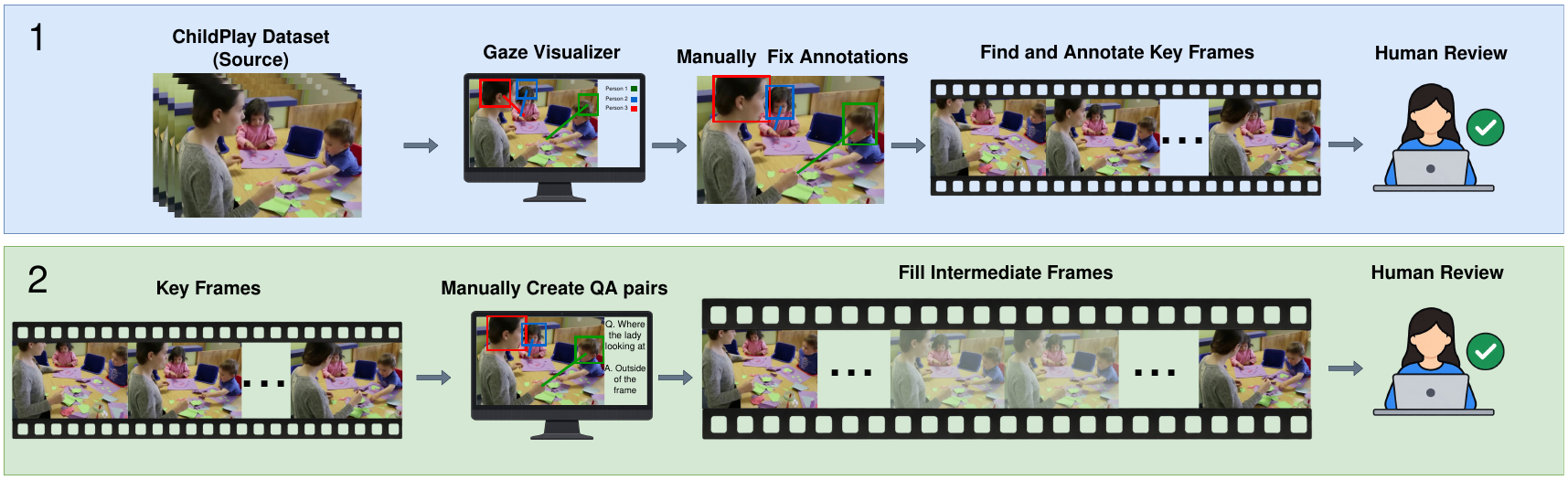}
    \caption{
    \textbf{\wgaze dataset construction and annotation pipeline.}
    In the first stage, the ChildPlay dataset is processed using our gaze visualiser, 
    incorrect annotations are manually corrected, representative key frames are identified 
    and annotated, and the resulting annotations undergo human review. In the second stage, 
    QA pairs are manually created for the verified key frames and propagated to the corresponding 
    intermediate frames, followed by a final human review to correct or rewrite inaccurate or 
    inconsistent QA pairs.
    }
    \label{fig:g3vqa_annotation}
\end{figure*}

\paragraph{Attended object} The second group concerns what people look at. In gaze object description, a person is identified by appearance and the question asks what that person looks at, so the VLLM must first find the correct person and then follow the gaze, rather than guess an object that is commonly looked at. In counting, the question asks how many people look at a given object, which requires the VLLM to verify where each arrow ends. In crowded videos with more than three people, the answer counts everyone who looks at the object, including people without a head box or an arrow, so the VLLM must also locate these people and determine where they look.
Two further types test for hallucination. Absent object questions refer to an object that is not in the scene or that nobody looks at, and since the correct answer is that nobody looks at it, a VLLM that reports a viewer is hallucinating one. Ambiguous description questions describe a person in a way that matches more than one person in the frame, and since the correct answer is that the description is not unique, a VLLM that selects one person and answers regardless fails to recognise the ambiguity.

\paragraph{Social gaze} The third group requires reasoning over the gaze of several people at once. Shared gaze asks whether people look at the same object, so the VLLM must combine the gaze of all people and report when no shared object exists. Exchanged gaze asks whether two people look at each other, which differs from looking at a common object and therefore requires verifying that the gaze of each person reaches the other. Engagement asks whether anyone looks away from the group, which requires the VLLM to follow every person at the same time.

\subsection{Temporal Questions}
Since frame-level questions refer to a single frame, they cannot test what a VLLM remembers, and we therefore add questions about whole clips, where the VLLM observes the clip as a video with the same arrows. These questions are of two kinds. Individual questions follow one person, for example when that person first looks at an object, how many times they look at it, and what they look at before or after it. Social questions follow the group, for example who looks at an object first, how many times two people look at each other, and whether and how many times one person blocks another person's face (Supp.\ \cref{sec:supp-temporal-examples}).
Answering these questions requires the VLLM to retain earlier events, preserve their order, and count separate looks instead of frames. Some questions also contain a false premise, such as a question about an object that nobody looks at, so a VLLM that always produces an answer is penalised. We further test memory through the way we treat hidden targets. When an object is hidden for a short time while the person keeps looking at the same place, we count it as a single look and the object retains its label, so the VLLM cannot rely on describing the visible frame and must remember the object.

\subsection{Dataset and Annotation}
\wgaze is built on 381 clips from ChildPlay~\cite{Tafasca_2023_ICCV}, comprising approximately 640K frame-level question-answer pairs and 16K temporal questions. \Cref{tab:splits} summarises their distribution across the training, validation, and test splits.
\begin{table}[h]
\centering
\caption{\textbf{Size of the \wgaze splits, rounded.}}
\label{tab:splits}
\scriptsize
\setlength{\tabcolsep}{4pt}
\renewcommand{\arraystretch}{1.08}
\resizebox{\columnwidth}{!}{%
\begin{tabular}{@{}lcccc@{}}
\toprule
\textbf{Split} & \textbf{Clips} & \textbf{Frames} & \textbf{Frame-level QA} & \textbf{Temporal QA} \\
\midrule
Train & 312 & 95K & 525K & 12K \\
Validation & 26 & 10K & 40K & 1K \\
Test & 43 & 10K & 75K & 3K \\
\midrule
Total & 381 & 116K & 640K & 16K \\
\bottomrule
\end{tabular}%
}
\end{table}

We built \wgaze in the five steps below, and \cref{fig:g3vqa_annotation} shows the first three.
\begin{enumerate}
    \setlength{\itemsep}{0pt}
    \item \textbf{Fixing the gaze labels:} We fixed wrong gaze points and head boxes in 50 ChildPlay videos.
    \item \textbf{Choosing key frames:} We picked around 7K frames where the gaze of a person changes and labelled every person in them.
    \item \textbf{Writing frame-level questions:} We turned these labels into question-answer pairs.
    \item \textbf{Writing temporal questions:} We made questions about whole clips and checked every answer by hand.
    \item \textbf{Adding target boxes:} We added a box around the object that each person looks at.
\end{enumerate}

\paragraph{Fixing the gaze labels} Annotators loaded the original ChildPlay videos and labels into our gaze visualiser and went through each video. In 50 videos they found wrong gaze points, wrong head boxes, or both, and fixed them.

\paragraph{Choosing key frames} Annotators picked a key frame at each moment where the gaze of a person changes in a clear way, and we checked the fixed labels and the key frames again. On each key frame, annotators described every person by their appearance, so that a question can point to one specific person. They also marked where each person looks and which object they look at, and flagged frames where the gaze is unclear or not visible, or where the eyes are closed.

\begin{figure*}[h]
    \centering
    \includegraphics[width=\textwidth]{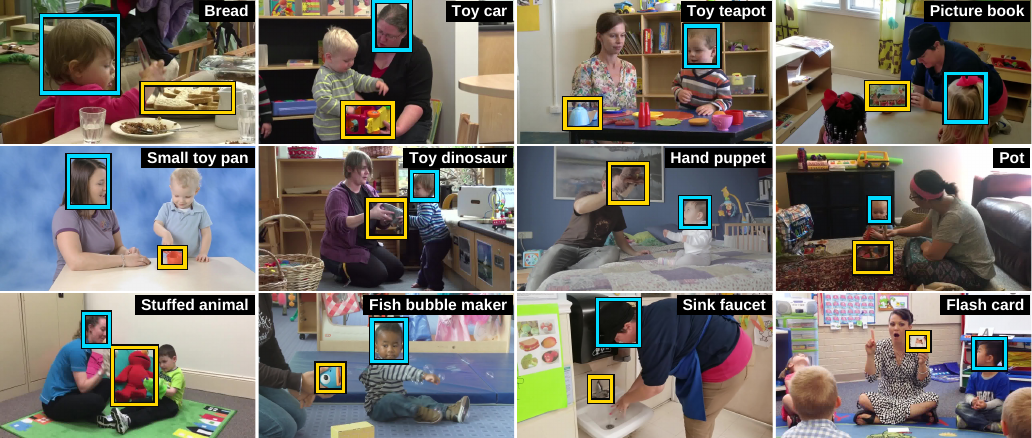}
    \caption{\textbf{Examples of gaze-target annotations across diverse social scenes.}
    Cyan boxes indicate the gazing person's head and yellow boxes mark the gaze target. Text labels describe the corresponding targets.}
    \label{fig:bbox_gaze}
\end{figure*}

\paragraph{Writing frame-level questions} The labels of a key frame stay valid until the next key frame. We used fixed templates to turn them into questions and answers about the gaze of each person and about social gaze in the scene, and annotators could edit or remove any pair. Annotators also wrote the ambiguous description questions and the counting answers for crowded videos with more than three people. For the absent object questions, they wrote or confirmed two lists for each video, one with objects that are not in the video and one with visible objects that nobody looks at, and we asked about each object of the first list on up to three key frames. The questions about a person were then carried over to the frames after a key frame until the next one, while the questions about the whole scene stayed on their key frame. At the end, all pairs were checked again, and wrong, unclear, or inconsistent ones were fixed or rewritten.

\paragraph{Writing temporal questions} The temporal questions come from our own labels, which are the object each person looks at, the gaze points and gaze types, the head boxes, and the appearance descriptions. For each clip, fixed templates turned these labels into questions with a proposed answer, and we saved the frames and people that support each answer. A frame counts as a look at an object only when the person has a usable gaze point and an object label. When the target is unknown for a few frames between two looks at the same object, the frames before and after decide whether this is one look or two. For the questions about blocked faces, we found events in the head boxes where two boxes overlap or one box disappears next to another person, and an annotator accepted or rejected each event and marked who blocks whom and in which frames. Finally, an annotator checked every temporal question against its supporting frames, kept it when the answer was right, fixed it when it was wrong, and removed it when it did not make sense for the clip.

\paragraph{Adding target boxes} On the key frames, annotators also added a box around the object that each person looks at (\cref{fig:bbox_gaze}). Each box started from the output of a grounding model, and annotators checked that the gaze point falls inside it and fixed it when it did not. The boxes of the frames between two key frames were copied or interpolated.

%% file: sec/4_experiments.tex
\section{Experiments}
\label{sec:experiments}

We evaluate \wgaze from three sides. First, we test whether our gaze tracking agent lowers the cost of video gaze estimation without losing accuracy, because it produces the gaze evidence that the models see. Second, we test the grounding of the attended object, which turns the predicted gaze into an object box. Finally, we benchmark a diverse set of vision large language models (VLLMs), ranging from 0.8B to 122B parameters.

\subsection{Evaluation Metrics}
\label{sec:eval_metric}

We evaluate gaze tracking using AUC, average L2 distance between predicted and annotated gaze points, and in/out AP, while measuring computational cost through estimator calls and FLOPs per call. For object grounding, we report mean intersection over union (mIoU), accuracy at IoU thresholds of 0.5 and 0.75 (Acc@50 and Acc@75), and average precision at IoU 0.5 (AP@50). Finally, VLLM responses are evaluated by accuracy, with yes/no, counting, and frame-based questions (28\% of the test questions) scored by exact match on the final answer. The remaining responses are assessed using AnyJev~\cite{zhang2026anyjev} (Supp.\ \cref{sec:supp-judge-prompt}), a single-forward-pass decision model that estimates answer correctness more efficiently than conventional LLM judges. Responses with a correctness probability of at least 0.5 are considered correct.

\subsection{Experimental Setup}
\label{sec:experimental_setup}

\begin{table*}[t]
\centering
\caption{\textbf{\wgaze benchmark results.}
    Accuracy by exact match or the AnyJev Tacit 9B judge. Individual, Group, and Ambiguous questions are about one person (one arrow), the scene (all arrows), or an ambiguous description (none). When, Count, and Memory ask for a frame, a number, and earlier targets, people, or events. Rules group models by size, and {\setlength{\fboxsep}{1pt}\colorbox{rank1}{Red}} marks the best per column in each group.}
\label{tab:w3_gaze_VQA_results}

\setlength{\tabcolsep}{4pt}
\renewcommand{\arraystretch}{1}
\setlength{\aboverulesep}{0pt}
\setlength{\belowrulesep}{0pt}

\resizebox{\textwidth}{!}{%
\begin{tabular}{l ccc ccc c}
\toprule
 & \multicolumn{3}{c}{\textbf{Frame-Level}} & \multicolumn{3}{c}{\textbf{Temporal}} & \\
\cmidrule(lr){2-4} \cmidrule(lr){5-7}
\multicolumn{1}{c}{\raisebox{13.7pt}[0pt][0pt]{\textbf{Model}}} & \shortstack{\textbf{Individual}\\(22,079)} & \shortstack{\textbf{Ambiguous}\\(2,863)} & \shortstack{\textbf{Group}\\(5,309)} & \shortstack{\textbf{When}\\(855)} & \shortstack{\textbf{Count}\\(627)} & \shortstack{\textbf{Memory}\\(1,235)} & \raisebox{9.6pt}[0pt][0pt]{\shortstack{\textbf{Overall}\\(32,968)}} \\
\midrule

Molmo2-4B
& 28.3\% & 11.5\% & 28.4\% & 27.3\% & 15.6\% & 16.5\% & 26.2\% \\

Qwen3.5-0.8B
& 27.2\% & 18.4\% & 32.6\% & \cellcolor{rank1}35.4\% & 8.0\% & 20.5\% & 26.9\% \\

InternVL3.5-2B
& 26.0\% & 31.3\% & 32.3\% & 31.1\% & 12.8\% & 20.2\% & 27.1\% \\

Molmo2-O-7B
& 28.2\% & \cellcolor{rank1}53.8\% & 29.1\% & 5.8\% & \cellcolor{rank1}25.8\% & 7.6\% & 29.2\% \\

Qwen3.5-2B
& \cellcolor{rank1}33.8\% & 17.1\% & 35.2\% & 31.3\% & 8.0\% & \cellcolor{rank1}22.2\% & 31.6\% \\

InternVL3.5-4B
& 30.5\% & 43.8\% & \cellcolor{rank1}38.0\% & 22.2\% & 13.6\% & 19.4\% & \cellcolor{rank1}31.9\% \\

\midrule

InternVL3.5-8B
& 22.2\% & 22.4\% & 40.1\% & 27.8\% & 13.2\% & 17.6\% & 24.9\% \\

InternVL3.5-GPT-OSS-20B-A4B
& 24.5\% & 20.6\% & 40.1\% & \cellcolor{rank1}33.1\% & 11.5\% & \cellcolor{rank1}21.6\% & 26.5\% \\

Molmo2-8B
& \cellcolor{rank1}30.9\% & 13.4\% & 28.2\% & 16.1\% & 12.1\% & 18.2\% & 27.7\% \\

InternVL3.5-14B
& 26.7\% & \cellcolor{rank1}51.2\% & \cellcolor{rank1}42.7\% & 26.3\% & \cellcolor{rank1}22.3\% & 19.0\% & \cellcolor{rank1}31.0\% \\

\midrule

InternVL3.5-30B-A3B
& 27.5\% & 39.7\% & 29.3\% & 29.2\% & 15.5\% & 19.7\% & 28.4\% \\

InternVL3.5-38B
& 30.0\% & 43.1\% & 35.4\% & \cellcolor{rank1}32.7\% & 18.8\% & 24.0\% & 31.7\% \\

Qwen3.6-35B-A3B
& 37.2\% & \cellcolor{rank1}67.2\% & 42.7\% & 25.5\% & 17.2\% & 22.4\% & 39.5\% \\

Qwen3.5-122B-A10B
& 38.4\% & 61.7\% & 42.8\% & 28.0\% & \cellcolor{rank1}20.4\% & \cellcolor{rank1}24.5\% & 40.0\% \\

Qwen3.5-27B
& 38.6\% & 65.2\% & 42.0\% & 27.3\% & 15.3\% & 22.7\% & 40.1\% \\

Qwen3.5-35B-A3B
& \cellcolor{rank1}39.1\% & 65.2\% & \cellcolor{rank1}42.9\% & 26.8\% & 19.9\% & 22.9\% & \cellcolor{rank1}40.7\% \\

\bottomrule

\end{tabular}%
}
\end{table*}

\paragraph{Gaze tracking}
Our gaze tracking agent extends an image-based gaze estimator to video by estimating gaze only when a new target has to be established and by following that target through the frames that come after. Once a target is found, the agent moves it from frame to frame with Lucas--Kanade optical flow, which avoids repeated estimation while the attention of a person stays on the same place. In parallel, OpenFace 3.0~\cite{hu2025openface30lightweightmultitask} monitors the head and eye behaviour of each person and calls the estimator again when it detects a substantial change, using an angular threshold of $2.0^{\circ}$. To stop tracking errors from accumulating, the estimator is also called again when the followed target moves more than one head diagonal away from its last confirmed position. We use PaGE as the estimator in all experiments (14 estimators in Supp.\ \cref{sec:supp-config}). \Cref{tab:gaze_tracking_comparison} shows that the agent then needs 85\% fewer estimator calls than running PaGE on every person in every frame, while its AUC and in/out AP are almost identical and its average distance error is only slightly higher.

\begin{table}[h]
\centering
\caption{\textbf{Gaze tracking accuracy and cost.}
Our agent maintains comparable accuracy
while reducing estimator calls by 85\%.
{\setlength{\fboxsep}{1pt}\colorbox{rank1}{Red}} marks
the best result,
{\setlength{\fboxsep}{1pt}\colorbox{rank2}{orange}}
the second, and
{\setlength{\fboxsep}{1pt}\colorbox{rank3}{yellow}}
the third.}
\label{tab:gaze_tracking_comparison}

\scriptsize
\setlength{\tabcolsep}{3pt}
\renewcommand{\arraystretch}{1.4}
\setlength{\aboverulesep}{0pt}
\setlength{\belowrulesep}{0pt}

\resizebox{\columnwidth}{!}{%
\begin{tabular}{lccccc}
\toprule
\textbf{Method} &
\textbf{Calls $\downarrow$} &
\textbf{FLOPs/Call} &
\textbf{AUC $\uparrow$} &
\textbf{Avg. L2 $\downarrow$} &
\textbf{In/out AP $\uparrow$} \\
\midrule
PaGE
& 25,017
& \cellcolor{rank2}0.293 T
& \cellcolor{rank1}0.9709
& \cellcolor{rank1}0.0730
& \cellcolor{rank1}0.9980 \\
Gaze-LLE
& 25,017
& \cellcolor{rank3}0.733 T
& \cellcolor{rank3}0.9493
& \cellcolor{rank3}0.1016
& \cellcolor{rank3}0.9916 \\
GazeAnywhere
& 25,017
& 0.922 T
& 0.9416
& 0.1190
& 0.9658 \\
Sharingan
& 25,017
& \cellcolor{rank1}0.048 T
& 0.9258
& 0.1134
& 0.9905 \\
Gupta et al.
& 25,017
& 1.392 T
& 0.9218
& 0.1378
& -- \\
Ours w/ PaGE
& \cellcolor{rank1}3,816
& \cellcolor{rank2}0.293 T
& \cellcolor{rank2}0.9705
& \cellcolor{rank2}0.0807
& \cellcolor{rank2}0.9979 \\
\bottomrule
\end{tabular}%
}
\end{table}

\paragraph{Gaze target grounding}
We ground attended objects using the gaze target agent (GTA)~\cite{nasiriboukani2026gaze}, which selects the detector box best aligned with the predicted gaze. This is challenging because gaze estimators provide points rather than object boxes, while detectors lack gaze information. Small gaze errors can therefore lead to incorrect matches, particularly for small or closely spaced objects. \Cref{tab:grounding_results} shows that RF-DETR 2XL achieves the highest accuracy among the three tested detectors, although box overlap remains moderate.

\begin{figure*}[t]
    \centering
    \IfFileExists{figures/failure.pdf}{\includegraphics[width=\textwidth]{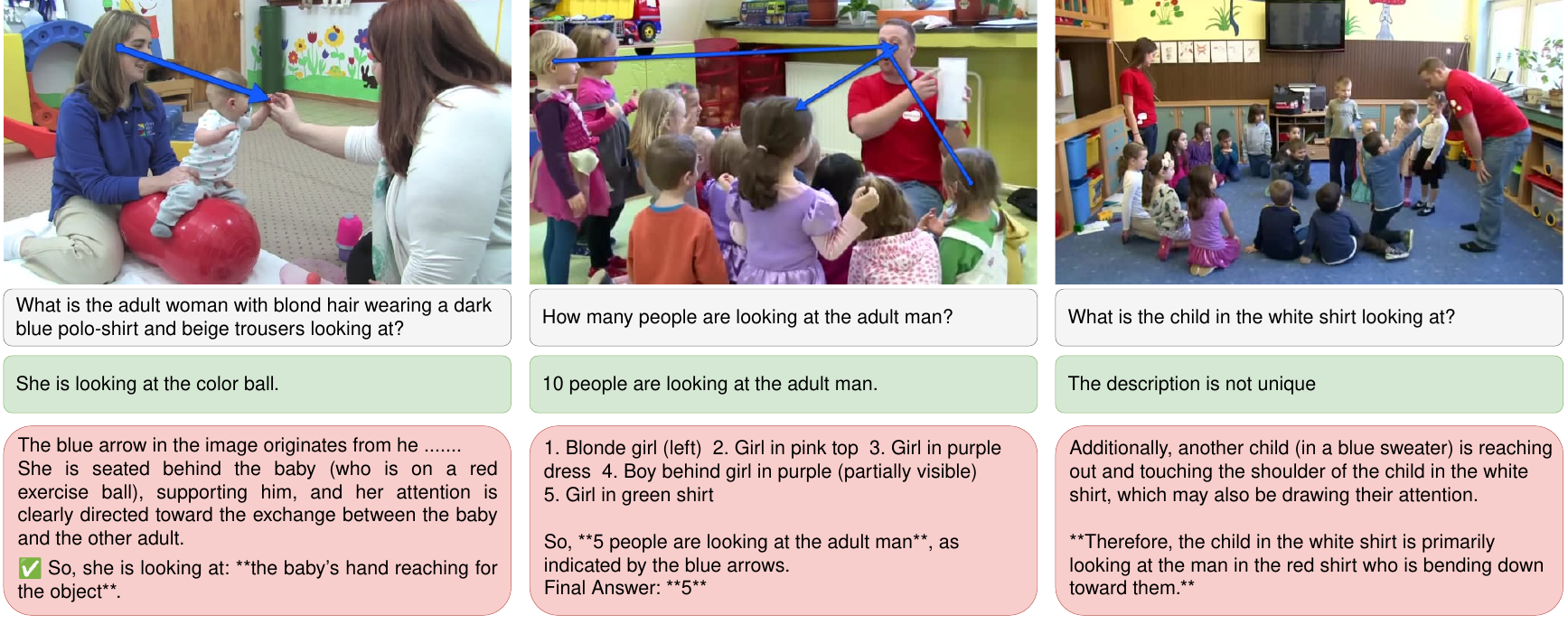}}{\fbox{\parbox{0.96\textwidth}{\centering\vspace{3.5cm}figures/failure.pdf\vspace{3.5cm}}}}
    \caption{\textbf{Failure cases of the best-performing model.}
    Even with accurate gaze arrows (none for the ambiguous question), most tested VLLMs fail on these examples. Green indicates ground truth and red shows Qwen3.5-35B-A3B predictions.
    }
    \label{fig:failure_cases}
\end{figure*}

\begin{table}[t]
\centering
\caption{\textbf{Gaze target grounding.}
Agreement between predicted and annotated gaze-target boxes
on the test set.
{\setlength{\fboxsep}{1pt}\colorbox{rank1}{Red}}
marks the best result per column.}
\label{tab:grounding_results}

\scriptsize
\setlength{\tabcolsep}{7pt}
\renewcommand{\arraystretch}{1.4}
\setlength{\aboverulesep}{0pt}
\setlength{\belowrulesep}{0pt}

\resizebox{\columnwidth}{!}{%
\begin{tabular}{lcccc}
\toprule
\textbf{Detector}
& \textbf{mIoU} $\uparrow$
& \textbf{Acc@50} $\uparrow$
& \textbf{Acc@75} $\uparrow$
& \textbf{AP@50} $\uparrow$ \\
\midrule
WeDetect-Large-Uni
& 27.9\% & 25.9\% & 22.5\% & 10.3\% \\
YOLO26x
& 29.4\% & 29.4\% & 23.9\% & 9.4\% \\
RF-DETR 2XL
& \cellcolor{rank1}31.8\%
& \cellcolor{rank1}32.0\%
& \cellcolor{rank1}26.0\%
& \cellcolor{rank1}11.6\% \\
\bottomrule
\end{tabular}%
}
\end{table}

\paragraph{Vision large language models}
We evaluate 16 VLLMs from the Qwen, InternVL, and Molmo families, ranging from 0.8B to 122B parameters, without additional training. Of the 75K frame-level test pairs, the tracker answers the 44K gaze point and direction questions (a larger share than in \cref{fig:g3vqa_tasks}) and 111 lack a predicted gaze point, so the VLLMs answer 30,251 frame-level and 2,717 temporal questions. Following the GGVL approach~\cite{nasiriboukani2025gaze}, each model receives the question, a brief explanation, and visual gaze prompts in which blue arrows connect people's heads to gaze points predicted by our PaGE-based agent. Individual questions show one arrow, group questions show all arrows, and ambiguous descriptions show none. Frame-level questions use a single image, while temporal questions use videos of up to 64 frames with visible frame numbers (prompts in Supp.\ \cref{sec:supp-prompts}).

\Cref{tab:w3_gaze_VQA_results} shows that \wgaze remains challenging for current VLLMs. Even the best-performing model, Qwen3.5-35B-A3B, achieves only 40.7\% overall accuracy. Average accuracy drops from 32\% on frame-level questions to 21\% on temporal questions, highlighting the difficulty of tracking visual attention across frames and recalling earlier gaze targets. Models also hallucinate people or objects and fail to recognise ambiguous descriptions. Moreover, larger models do not consistently outperform smaller ones on temporal reasoning tasks.

\raggedbottom
\subsection{Failure Analysis}
\label{sec:analysis}

The accuracy of the VLLMs depends strongly on the type of question, and \cref{fig:failure_cases} shows three situations in which most models fail even though the gaze arrows are accurate, so the errors come from the VLLM and not from the gaze estimator. When a person looks at a small object, the models often name the person or the surroundings next to the arrow instead of the object at its tip, and even when the arrow ends inside the attended object, the best model answers only about half of the questions about one person correctly. In crowded scenes the correct count includes everyone who looks at the object, but arrows are drawn only for people with a head box and the models mostly report a count close to the number of arrows, so the best model answers only 10\% of the questions with at least four viewers. The ambiguous descriptions show the same shortcut, because ten of the 16 models miss the ambiguity in most of these questions and describe one of the matching people. The temporal questions are the hardest, since the models count frames instead of separate looks and have trouble remembering what a person looked at earlier, and a more accurate arrow does not help here. In one video, for example, a woman looks at the hand of a girl three times, but the best model answers 100, which counts frames instead of looks. In each case the VLLM answers with what is easiest to see, namely the nearest person, the visible arrows, a single person, or the number of frames, instead of reasoning about who looks at what and when.

%% file: sec/6_conclusion.tex
\section{Conclusion}
\label{sec:conclusion}

We introduced \wgaze, a manually annotated video benchmark for social and temporal gaze reasoning with around 640K frame-level questions in nine types and around 16K temporal questions in 21 types over 381 ChildPlay clips, together with corrected gaze labels for 50 videos. We also introduced a training-free gaze tracking agent that needs 85\% fewer estimator calls than estimating gaze in every frame with almost the same accuracy. Across 16 VLLMs, even the best model answers only 41\% of the questions correctly, and the average accuracy falls from 32\% on frame-level questions to 21\% on temporal questions. Failure cases show that the VLLMs, not the gaze estimator, cause these errors, since the models answer with what is easiest to see, such as the nearest person, the visible arrows, or the number of frames.

\paragraph{Future work} \wgaze covers the whole chain from gaze estimation and tracking to object grounding and question answering, and we evaluate each part separately. A natural next step is an agent in which a VLLM calls our gaze tracker and a grounding model as tools and answers from their output. Such an agent could also keep the record of where each person looked over time, which our tracker already produces, so that the VLLM can look up earlier looks instead of counting frames. Since the detectors find small toys and cards poorly, better grounding of small objects would make the attended object boxes more reliable. Finally, all VLLMs were tested without training, and we hope that the large training split of \wgaze will help to develop VLLMs that follow attention through time and understand who looks at what.

\section*{LLM Usage Statement}
Claude Opus 5.5 (Anthropic) was used in the preparation of this manuscript solely to assist with language editing,
including improving wording, grammar, clarity, and readability, as well as providing support with manuscript and
figure formatting. It was not used to create or verify the benchmark annotations, design experiments, analyse
experimental results, or draw scientific conclusions. All technical content, annotations, methodological decisions,
experimental design, implementation, analysis, and conclusions were produced and verified by the authors, who take full responsibility for the content of this work.